\documentclass[pdflatex,sn-basic,iicol,sort&compress]{sn-jnl}%

\usepackage{graphicx}
\usepackage{wrapfig,lipsum}
\usepackage{amssymb}
\usepackage{amsmath}
\usepackage{afterpage}
\usepackage{xcolor}
\usepackage{booktabs}
\usepackage{fontawesome}
\setcitestyle{numbers,square,citesep={,}} 

\usepackage{multicol}
\usepackage{ulem}

\definecolor{rosso}{cmyk}{0,1,1,0.4}
\definecolor{rossos}{cmyk}{0,1,1,0.6}
\definecolor{rossoc}{cmyk}{0,1,1,0.2}
\definecolor{blu}{cmyk}{1,1,0,0.3}
\definecolor{blus}{cmyk}{1,1,0,0.3}
\definecolor{bluc}{cmyk}{1,1,0,0.1}
\definecolor{verde}{cmyk}{0.92,0,0.59,0.25}
\definecolor{verdec}{cmyk}{0.92,0,0.59,0.15}
\definecolor{verdes}{cmyk}{0.92,0,0.59,0.4}
\hypersetup{colorlinks,bookmarksopen,bookmarksnumbered,
linkcolor=blus,pdfstartview=FitH,urlcolor=rossos,citecolor=verdes}

\newcommand{\be}{\begin{equation}}
\newcommand{\ee}{\end{equation}} 
\newcommand{\bry}{\begin{array}}
\newcommand{\ery}{\end{array}} 
\newcommand{\dst}{\displaystyle} 
\newcommand{\bit}{\begin{itemize}} 
\newcommand{\eit}{\end{itemize}} 
\newcommand{\ben}{\begin{enumerate}} 
\newcommand{\een}{\end{enumerate}}

\DeclareMathOperator*{\argmax}{arg\,max}

\begin{document}

\title[Comparison of Affine and Rational Quadratic Spline Coupling and Autoregressive Flows through Robust Statistical Tests]{\centering Comparison of Affine and Rational Quadratic Spline Coupling and Autoregressive Flows through Robust Statistical Tests}

\author*[1]{\fnm{Andrea} \sur{Coccaro}}\email{andrea.coccaro@ge.infn.it}
\author*[1,2]{\fnm{Marco} \sur{Letizia}}\email{marco.letizia@edu.unige.it}
\author*[1,3,4]{\fnm{Humberto} \sur{Reyes-Gonz\'{a}lez}}\email{humberto.reyes@rwth-aachen.de}
\author*[1]{\fnm{Riccardo} \sur{Torre}}\email{riccardo.torre@ge.infn.it}

\affil*[1]{\orgdiv{INFN}, \orgname{Sezione di Genova}, \orgaddress{\street{Via Dodecaneso 33}, \city{Genova}, \postcode{16146}, \country{Italy}}}
\affil*[2]{\orgdiv{MaLGa - DIBRIS}, \orgname{University of Genova}, \orgaddress{\street{Via Dodecaneso 35}, \city{Genova}, \postcode{16146}, \country{Italy}}}
\affil*[3]{\orgdiv{Department of Physics}, \orgname{University of Genova}, \orgaddress{\street{Via Dodecaneso 33}, \city{Genova}, \postcode{16146}, \country{Italy}}}
\affil*[4]{Institut f\"{u}r Theoretische Teilchenphysik und Kosmologie, \orgdiv{RWTH Aachen}, \city{Aachen}, \postcode{52074}, \country{Germany}}

\abstract{Normalizing flows have emerged as a powerful brand of generative models, as they not only allow for efficient sampling of complicated target distributions but also deliver density estimation by construction. We propose here an in-depth comparison of coupling and autoregressive flows, both based on symmetric (affine) and non-symmetric (rational quadratic spline) bijectors, considering four different architectures: real-valued non-Volume preserving (RealNVP), masked autoregressive flow (MAF), coupling rational quadratic spline (C-RQS), and autoregressive rational quadratic spline (A-RQS). We focus on a set of multimodal target distributions of increasing  dimensionality ranging from 4 to 400. The performances were compared by means of different test statistics for two-sample tests, built from known distance measures: the sliced Wasserstein distance, the dimension-averaged one-dimensional Kolmogorov--Smirnov test, and the Frobenius norm of the difference between correlation matrices. Furthermore, we included estimations of the variance of both the metrics and the trained models. Our results indicate that the A-RQS algorithm stands out both in terms of accuracy and training speed. Nonetheless, all the algorithms are generally able, without too much fine-tuning, to learn complicated distributions with limited training data and in a reasonable time of the order of hours on a Tesla A40 GPU. The only exception is the C-RQS, which takes significantly longer to train, does not always provide good accuracy, and becomes unstable for large dimensionalities. All algorithms were implemented using \textsc{TensorFlow2} and \textsc{TensorFlow Probability} and have been made available on \textsc{GitHub} \href{https://github.com/NF4HEP/NormalizingFlowsHD}{\faGithub}.
}


\maketitle
\clearpage\markboth{}{}

\tableofcontents

\section{Introduction}
The modern data science revolution has opened a great window of opportunities for scientific and societal advancement. In particular, Machine Learning (ML) technologies are being applied in a wide variety of fields from finance to astrophysics. It is thus crucial to carefully study the capabilities and limitations of ML methods in order to ensure their systematic usage.
This is particularly pressing when applying ML to scientific research, for instance in a field such as High Energy Physics (HEP), where one often deals with complicated high-dimensional data and high levels of precision are needed. 

In this paper, we focus on Normalizing Flows (NFs) \cite{NFs1,NFs2,https://doi.org/10.48550/arxiv.1505.05770,dinh2015nice}, a~class of neural density estimators that for one, offers a competitive approach to generative models, such as Generative Adversarial Networks (GANs) \cite{Goodfellow2014} and Variational AutoEncoders (VAEs) \cite{Kingma2014,pmlr-v32-rezende14}, for~the generation of synthetic data and, for another, opens up a wide range of applications due to its ability to directly perform density estimation. Even though we have in mind applications of NFs to HEP, in this paper, we remain agnostic with respect to the applications and only performed a general comparative study of the performances of coupling and autoregressive NFs when used to learn high-dimensional multi-modal target distributions. Nevertheless, it is worth mentioning some of the applications of NFs to HEP can also be extended to several other fields of scientific research. 

While applications of the generative direction of NFs is rather obvious in a field such as HEP, which bases its foundations on Monte Carlo simulations, it is interesting to mention some of the possible density estimation applications. The~ability to directly learn the likelihood, or the posterior in a Bayesian framework, has applications ranging from analysis, inference, reinterpretation, and preservation to simulation-based likelihood-free inference \cite{Fan2013,Papamakarios2016,Brehmer:2018hga,Green:2020hst,Villar:2022are,Campagne:2023ter}, unfolding of HEP analyses \cite{Bellagente:2022jcj}, generation of effective priors for Bayesian inference \cite{Zoran2011FromLM,Green:2020dnx,Glusenkamp:2020gtr,KodiRamanah:2020nbc,Cheung:2021orb,Ruhe:2022ddi}, systematic uncertainty estimation and parametrization, generation of effective proposals for sequential Monte Carlo \cite{Gu2015,10.5555/3045390.3045710,Foreman:2021ljl,Hackett:2021idh,Singha:2023cql,Caselle:2022esc,Matthews:2022sds,Cranmer:2023xbe}, numerical integration based on importance sampling algorithms \cite{8c14e8dabdfe4a7da84cb401a48533f3,Gabrie:2021tlu,Pina-Otey:2020hzm,Gao:2020zvv}, and probabilistic programming applied to fast inference \cite{pmlr-v54-le17a,10.5555/3546258.3546315}.

The basic principle behind NFs is to perform a series of invertible bijective transformations on a simple base probability density function (PDF) to approximate a complicated PDF of interest. The optimal parameters of the transformations, often called ``bijectors'', are derived from training neural networks (NNs) that directly take the negative log-likelihood of the true data computed with the NF distribution as the loss function. As it turns out, PDFs are everywhere in HEP: from the likelihood function of an experimental or a phenomenological result to the distribution that describes a particle-collision process. 
Thus, NFs have found numerous applications in HEP: they have been used for numerical integration and event generation \cite{Butter:2021csz, Verheyen:2022tov, Krause:2021ilc, Krause:2021wez, Gao:2020vdv, Heimel:2022wyj,Heimel:2023ngj,Ernst:2023qvn}, anomaly detection \cite{Nachman:2020lpy,Golling:2022nkl,10.21468/SciPostPhys.12.2.077}, detector unfolding \cite{Bellagente:2020piv,Backes:2022vmn}, etc.

The growing interest in NFs implies the urgency of testing state-of-the-art architectures against complex data to ensure their systematic usability and to assess their expected performances. The purpose of this work was then to evaluate the performance of NFs against generic complicated distributions of increasing dimensionality. By performing this study, we aimed to make a concrete step forward in the general understanding of the realistic performances and properties of NFs, especially in high-precision scenarios. This work comprises a substantial upgrade with respect to our early study of Ref.~\cite{Reyes-Gonzalez:2022rco}, as we now have included more NF architectures, extended the dimensionality of the distributions, and significantly improved the testing strategy.

Our strategy was the following. We implemented in \textsc{Python}, using \textsc{TensorFlow2} with \textsc{TensorFlow Probability}, four of the most commonly used NF architectures of the coupling and autoregressive type: Real-Valued Non-Volume Preserving (RealNVP) \cite{dinh2017density}, Masked Autoregressive Flow (MAF) \cite{papamakarios2018masked}, Coupling Rational Quadratic Spline (C-RQS) \cite{durkan2019neural}, and Autoregressive Rational Quadratic Spline (A-RQS) \cite{durkan2019neural}. 

We tested these NF architectures considering Correlated Mixture of Gaussian (CMoG) multi-modal distributions with dimensionalities ranging from 4 to 400. We also performed a small-scale hyperparameter scan, explicitly avoiding the fine-tuning of the models and generating the best result for each NF architecture and target distribution. 

The performances were measured by means of different test statistics for two-sample testing built from known distance measures: the sliced Wasserstein distance, the dimension-averaged one-dimensional Kolmogorov--Smirnov statistic, and the Frobenius norm of the difference between correlation matrices. {The analyses were performed by comparing a test sample, drawn from the original distribution, with an NF-generated one.} Moreover, all test-statistics calculations were cross-validated, and an error was assigned both to the evaluation procedure, with repeated calculations of the metrics on different instances of the test and NF-generated samples, and to the training procedure, with repeated calculations on models trained with different instances of the training~sample.

This paper is organized as follows. In Section \ref{Sec:NFs}, we describe the concept of NFs in more detail, focusing on the coupling and autoregressive types. In Section \ref{Sec:NFimps}, we introduce the specific NF architectures under investigation. In Section \ref{sec:metrics}, we present the metrics used in our analysis, and in Section \ref{Sec:Tests}, we discuss our results. Finally, we provide our concluding remarks in Section \ref{ref:conclusion}, with emphasis on the several prospective research avenues that we plan to follow.

\section{Normalizing Flows}\label{Sec:NFs}

Normalizing Flows are made of series of bijective, continuous, and invertible transformations that map a simple \textit{base}
 PDF to a more complicated \textit{target} PDF. The purpose of NFs is to estimate the unknown underlying distribution of some data of interest and to allow for the generation of samples approximately following the same distribution. Since the parameters of both the base distribution and the transformations are known, one can \textit{generate} samples from the target distribution by drawing samples from the base distribution and then applying the proper transformation. This is known as the \textit{generative direction} of the flow. Furthermore, since the NF transformations are invertible, one can also obtain the probability density of the true samples via inverse transformations from the target to the base PDF. This is known as the \textit{normalizing direction} of the flow. It is called ``normalizing'' because the base distribution is often Gaussian even though this is not a requirement, and this is also the origin of the name ``normalizing flows''.

The basic idea behind NFs is the change of variable formula for a PDF. Let $X, Y \in \mathbb{R}^{D}$ be random variables with PDFs $p_X, p_Y:  \mathbb{R}^{D} \rightarrow \mathbb{R}$.  Let us define a bijective map $\mathbf{g}:X\rightarrow Y$, with inverse $\mathbf{f}=\mathbf{g}^{-1}$. The two densities are then related by the well known formula
 
\begin{equation}\label{NF1}
\bry{lll}
p_Y(y) &=& p_X(\mathbf{g}^{-1}(y))\mid\det \mathrm{J}_{g}\mid^{-1}\, \vspace{2mm}\\
&=&p_X(\mathbf{f}(y))\mid\det\mathrm{J}_{f}\mid \,,
\ery
\end{equation}
where $\mathrm{J}_{f}=\frac{\partial \mathbf{f}}{\partial y}$ is the Jacobian of $\mathbf{f}(y)$, and $\mathrm{J}_{g}=\frac{\partial \mathbf{g}}{\partial x}$ is the Jacobian of $\mathbf{g}(x)$.\footnote{Throughout the paper we always interpret $X$ as the base distribution and $Y$ as the target distribution, i.e., the data. We also always model flows in the generative direction, from base to data.}

Let us now consider a set of parameters $\{\phi\}$ characterizing the chosen base density $p_{X}$ (typically the mean vector and covariance matrix of a multivariate Gaussian) and parametrize the map $\bf g$ by another set of parameters $\{\theta\}$. One can then perform a maximum likelihood estimation of the parameters $\Phi=\{\phi,\theta\}$ given some measured data $\mathcal{D}=\lbrace y^{I}\rbrace^{N}_{I=1}$ distributed according to the unknown PDF $p_{y}$. The log-likelihood of the data is given by the following expression:

\begin{equation}\label{loss}
\bry{lll}
\log p(\mathcal{D}\mid\Phi)&=&\dst \sum_{I=1}^{N}\log p_{Y}(y^{I}\mid\Phi) \vspace{2mm}\\
&\hspace{-1.8cm}=&\dst\hspace{-1.8cm}\sum^{N}_{I=1}\log p_{X}(\mathbf{f}_{\theta}(y^{I})\mid\theta,\phi)+\log\mid\det\mathrm{J}_{f}\mid\,
\ery
\end{equation}
where we make the dependence of $\mathbf{f}$ on $\theta$ explicit through the notation $\mathbf{f}_{\theta}$. Then, the best estimate of the parameters $\Phi$ is given by

\begin{equation}
\hat{\Phi}=\argmax_{\Phi}\log p(\mathcal{D}\mid\Phi)\,
\end{equation}
Once the parameters $\hat{\Phi}$ have been estimated from the data, the approximated target distribution can be sampled by applying the generative map $\bf g$ to samples obtained from the base PDF. The normalizing direction $\bf f$ can instead be used to perform density evaluation by transforming the new data of interest into sample generated by the base PDF, which is easier to evaluate.

Beside being invertible, the map $\bf g$ should satisfy the following properties:
\begin{itemize}
\item It should be sufficiently expressive to appropriately model the target distribution;
\item It should be computationally efficient, meaning that both $\mathbf{f}$ (for training, this means computing the likelihood) and $\mathbf{g}$ (for generating samples), as well as their Jacobian determinants, must be easily calculable.
\end{itemize}

The composition of invertible bijective functions is also an invertible bijective function. Thus, $\mathbf{g}$ can be generalized to a set of $N_{t}$ transformations as $\mathbf{g}=g_{N_{t}}\circ g_{N_{t}-1}\circ ...g_1$ with inverse $\mathbf{f}=f_1\circ ... f_{N_{t}-1} \circ f_{N_{t}}$ and $\det \mathrm{J}_{f}=\prod_{n=1}^{N_{t}}\det \mathrm{J} _{f_n}$, where each $f_n = g_{n}^{-1}$ depends on a  $y_n$ intermediate random variable. This is a standard strategy to increase the flexibility of the overall~transformation.

Typically, but not mandatorily, NF models are implemented using NNs to determine the parameters of the bijectors. The optimal values are obtained by minimizing a loss function corresponding to minus the log-likelihood defined as in Eq.~\eqref{loss}. This is a natural approach when the samples from the target density are available but the density itself cannot be evaluated, a common occurrence in fields, such as HEP, that heavily rely on Monte Carlo simulations.\footnote{Approaches beyond maximum likelihood, which use different loss functions, have also been considered in the literature, such as in \mbox{Refs \cite{No2018BoltzmannGS,Midgley2022FlowAI,Grover2017FlowGANCM,villani2003topics,Arjovsky2017WassersteinGA,Tolstikhin2017WassersteinA}}. Inthis paper we always use the maximum likelihood approach and minus the log-likelihood as loss function.} This makes the models extremely flexible, with a usually stable training, at the cost of a potentially large number of parameters. Nonetheless, the flow transformation must be carefully designed; for instance, even if a given map and its inverse, with~their respective Jacobians, are computable, one direction might be more efficient than the other, leading to models that favor sampling over evaluation (and training) or vice versa. Among the wide and growing variety of NF architectures available (see Ref.~\cite{9089305} for an overview), we focus in this work on {\it coupling} \cite{dinh2015nice} and {\it autoregressive} flows \cite{10.5555/3157382.3157627}, arguably the most widely used implementations of NFs, particularly in HEP.

\subsection{Coupling Flows} 
Coupling flows, originally introduced in Ref.~\cite{dinh2015nice}, are made of stacks of so-called coupling layers, in~which each sample with dimension $D$ is partitioned into two samples $A$ and $B$ with dimensions $d$ and $D - d$, respectively. The~parameters of the bjiector transforming the sample $A$ are modeled by a NN that uses $B$ as input, effectively constructing the $p(y_{d} \vert x_{d-D})$ conditional probability distributions. At~each coupling layer in the stack, different partitionings are applied, usually by shuffling the dimensions before partitioning, so that all dimensions are properly~transformed.

In other words, starting from a disjoint partition of a random variable $Y\in  \mathbb{R}^{D} $ such that $(y^A,y^B)\in  \mathbb{R}^{d}\times  \mathbb{R}^{D-d}$ and a bijector $\mathbf{h}(\,\cdot\,;\theta):  \mathbb{R}^{d}\rightarrow  \mathbb{R}^{d}$, a~coupling layer maps $\mathbf{g}:X\rightarrow Y$ as follows:
\be
\bry{llll}
&y^{A}&=&\mathbf{h}(x^A;\Theta(x^B))\,,\vspace{2mm}\\
&y^{B}&=&x^B\,
\ery
\ee
where the parameters $\theta$ are defined by a generic function $\Theta(x^B)$ only defined on the $ \mathbb{R}^{D-d}$ partition, generally modeled by an NN. The~function $\Theta(x^B)$ is called a {\it conditioner}, while the bijectors $\mathbf{h}$ and $\mathbf{g}$ are called {\it coupling function} and {\it coupling flow}, respectively. The~necessary and sufficient condition for the coupling flow $\mathbf{g}$ to be invertible is that the coupling function $\mathbf{h}$ is invertible. In~this case, the inverse transformation is given by
\be
\bry{llll}
&x^{A}=\mathbf{h}^{-1}(y^A;\Theta(x^B))\,,\vspace{2mm}\\
&x^{B}=y^B\
\ery
\ee

Notice that despite the presence of a NN, whose inverse is unknown, to~parametrize the conditioner, the~invertibility of $\mathbf{h}$ is guaranteed by the fact that such a conditioner is a function of the unchanged dimensions only. The~Jacobian of $\mathbf{g}$ is then a two-block triangular matrix.
The dimensions $\lbrace 1:d \rbrace$ are given by the Jacobian of $\mathbf{h}$, and~the dimensions $\lbrace d:D\rbrace$ is the identity matrix. Thus, the~Jacobian determinant is simply the following:
\begin{equation}
\det \mathrm{J}_g=\prod_{i=1}^{d} \frac{\partial h_i}{\partial x^{A}_i}.
\end{equation}

Note that the choice of the partition is arbitrary. The~most common choice is to split the dimensions in half, but~other partitions are possible~\cite{9089305}. Obviously, even when dimensions are halved, the~way in which they are halved is not unique and can be implemented in several different ways, for~instance through random masks, or~through random shuffling before dividing the first and last~half.

\subsection{Autoregressive~Flows}
Autoregressive flows, first introduced in Ref.~\cite{10.5555/3157382.3157627}, can be viewed as a generalization of coupling flows. Now, the~transformations of each dimension $i$ are modeled by an autoregressive DNN according to the previously transformed dimensions of the distribution, resulting in the $p( y_i  \vert y_{1:i-1})$ conditional probability distributions, where $y_{1:i-1}$ is a shorthand notation to indicate the list of variables $y_{1},\ldots y_{i-1}$. After~each autoregressive layer, the~dimensions are permuted to~ensure the expressivity of the bijections over the full dimensionality of the target~distribution. 

Let us consider a bijector $\mathbf{h}(\,\cdot\,;\theta):  \mathbb{R}\rightarrow  \mathbb{R}$, parametrized by $\theta$. We can define an autoregressive flow function $\mathbf{g}$ such that

\be\label{eq:autoreg1}
\bry{llll}
&y_1&=&x_1\,,  \vspace{2mm}  \\ 
&y_i&=&\mathbf{h}(x_i;\Theta_{i}(y_{1:i-1}))\,,\quad i=2,\ldots,D\,
\ery
\ee

The resulting Jacobian of $\mathbf{g}$ is again a triangular matrix, whose determinant is easily computed as

\begin{equation}
\det \mathrm{J}_g=\prod_{i=1}^{D} \frac{\partial h_i}{\partial x_i}.
\end{equation}
where $\partial h_i/\partial x_i $ are the diagonal terms of the~Jacobian.

Given that the structure of the bijector is similar to that of the coupling flow, also, in this case, the bijector is referred to as a {\it coupling function}. Note that $\Theta_{j}$ can also be alternatively determined with the precedent untransformed dimensions of $X$ \cite{10.5555/3157382.3157627} such that

\be\label{eq:autoreg2}
\bry{llll}
&y_1&=&x_1\,,  \vspace{2mm}  \\ 
&y_i&=&\mathbf{h}(x_i;\Theta_{i}(x_{1:i-1}))\,,\quad i=2,\ldots,D\,
\ery
\ee

The choice of variables used to model the conditioner may depend on whether the NF is intended for sampling or density estimation. In~the former case, $\Theta$ is usually chosen to be modeled from the base variable $X$ so that the transformations in the generative direction would only require one forward pass through the flow. The~transformations in the normalizing direction would instead require $D$ iterations trough the autoregressive architecture. This case is referred to as {\it inverse autoregressive flow} \cite{10.5555/3157382.3157627}\footnote{Notice that in Ref.~\cite{9089305}, parametrizing the flow in the normalizing direction (the opposite of our choice), apparently uses the inverse of our formulas for direct and inverse flows. Our notation (and nomenclature) is consistent with that of Ref.~\cite{papamakarios2018masked}.} and corresponds to the transformations in Eq.~\eqref{eq:autoreg2}. Conversely, in~the case of density estimation, it is convenient to parametrize the conditioner using the target variable $Y$ since transformations would be primarily in the normalizing direction. This case is referred to as {\it direct autoregressive flow} and corresponds to the transformations in Eq.~\eqref{eq:autoreg1}. In~any case, when training the NFs, one always needs to perform the normalizing transformations to estimate the log-likelihood of the data, as~in Eq.~\eqref{loss}. In~our study, we only consider the direct autoregressive flow described by Eq.~\eqref{eq:autoreg1}.

\section{Architectures}\label{Sec:NFimps}
In the previous section, we describe NFs, focusing on the two most common choices for parametrizing the bijector $\mathbf{g}$ in terms of the coupling function $\mathbf{h}$. The~only missing ingredient to make NFs concrete, remains the explicit choice of $\mathbf{h}$. For~this study, we have chosen four of the most popular implementations of coupling and autoregressive flows: the real-valued non-volume preserving (RealNVP) \cite{dinh2017density}, the~masked autoregressive flow (MAF) \cite{papamakarios2018masked}, and~the coupling and autoregressive rational-quadratic neural spline flows (C-RQS and A-RQS) \cite{durkan2019neural}.\footnote{Reference~\cite{durkan2019neural} refers to coupling and autoregressive RQS flows as RQ-NSF (C) and RQ-NSF (AR), where RQ-NSF stands for rational-quadratic neural spline flow, and~A and C for autoregressive and coupling, respectively.} We discuss them in turn in the following subsections and give additional details about our specific implementation in Appendices \ref{app:RealNVP}--\ref{app:ARQS}.

\subsection{The~RealNVP} 
The RealNVP~\cite{dinh2017density} is a type of coupling flow whose coupling functions $\mathbf{h}$ are affine functions with the following form:

\small\be\label{eq:realnvp}
\bry{lllll}
&y_{i}&=&x_{i}\,,\quad &i=1,\ldots,d\,, \vspace{2mm} \\
&y_{i}&=&x_{i} e^{s_{i-d}(x_{1:d})}+t_{i-d}(x_{1:d})\,,\,\, &i=d+1,\ldots,D\,
\ery
\ee\normalsize
where the $s$ and $t$ functions, defined on $\mathbb{R}^{d}\to\mathbb{R}^{D-d}$, respectively correspond to the scale and translation transformations modeled by NNs. The~product in Eq.s \eqref{eq:realnvp} is intended elementwise for each $i$ so that, $x_{d+1}$ is multiplied by $s_{1}$, $x_{d+2}$ by $s_{2}$ and~so on up~to $x_{D}$, which is multiplied by $s_{D-d}$.
The Jacobian of this transformation is a triangular matrix with diagonal 
${\rm diag}(\mathbb{I}_d,{\rm diag}(\exp (s_{i-d}(x_{1:d}))))$ with $i=d+1,\ldots,D$, 
so that its determinant is independent of $t$ and is simply given by

\be\label{eq:realnvpjacobian}
\det \mathrm{J}=\prod_{i=1}^{D-d}e^{s_{i}(x_{1:d})}=\exp\left(\displaystyle\sum_{i=1}^{D-d}s_{i}(x_{1:d})\right)\,
\ee

The inverse of Eq.~\eqref{eq:realnvp} is given by
\small\be\label{eq:realnvpinverse}
\bry{lllll}
&\hspace{-2mm}x_{i}&\hspace{-1mm}=&\hspace{-1mm}y_{i}\,,\quad &\hspace{-1mm}i=1,\ldots,d\,,  \vspace{2mm}\\
&\hspace{-2mm}x_{i}&\hspace{-1mm}=&\hspace{-1mm}\dst\left(y_{i}-t_{i-d}(y_{1:d})\right) e^{-s_{i-d}(y_{1:d})}\,,\,\, &\hspace{-1mm}i=d+1,\ldots,D\,
\ery
\ee\normalsize

A crucial property of the affine transformation \eqref{eq:realnvp} is that its inverse \eqref{eq:realnvpinverse} is again an affine transformation depending only on $s$ and $t$, and~not on their inverse. This implies that the $s$ and $t$ functions can be arbitrarily complicated (indeed they are parametrized by a DNN), still leaving the RealNVP equally efficient in the forward (generative) and backward (normalizing) directions.

\subsection{The~MAF}
The MAF algorithm was developed starting from the masked autoencoder for distribution estimation (MADE) \cite{MADE2015} approach for implementing an autoregressive neural network through layers masking (see Appendix \ref{app:MAF}).

In the original MAF implementation~\cite{papamakarios2018masked}, the~bijectors are again affine functions \mbox{described as}
\small\be\label{eq:MAF1}
\bry{llll}
&y_{1}&=&x_{1}\,,\vspace{2mm} \\
&y_{i}&=&x_{i} e^{s_{i-1}(y_{1:i-1})}+t_{i-1}(y_{1:i-1})\,,\quad i=2,\ldots,D\
\ery
\ee\normalsize

The functions $s$ and $t$ are now defined on $\mathbb{R}^{D-1}\to\mathbb{R}^{D-1}$.
The determinant of the Jacobian is simply 
\be\label{eq:MAF1Jacobian}
\det \mathrm{J}=\prod_{i=1}^{D-1}e^{s_{i}(y_{1:i})}=\exp\left(\displaystyle\sum_{i=1}^{D-1}s_{i}(y_{1:i})\right)
\ee
and the inverse transformation is

\small\be\label{eq:MAF1inverse}
\bry{llll}
&\hspace{-2mm}x_{1}&\hspace{-1mm}=&\hspace{-1mm}y_{1}\,, \vspace{2mm}  \\
&\hspace{-2mm}x_{i}&\hspace{-1mm}=& \hspace{-1mm}\left(y_{i}-t_{i-1}(y_{1:i-1})\right) e^{-s_{i-1}(y_{1:i-1})}\,,\,\, i=2,\ldots,D\,.
\ery
\ee\normalsize

As in the case of the RealNVP, the~affine transformation guarantees that the inverse transformation only depends on $s$ and $t$ and not on their inverse, allowing for the choice of arbitrarily complicated functions without affecting computational~efficiency.

\subsection{The RQS~Bijector} 
The bijectors in a coupling or masked autoregressive flow are not restricted to affine functions. It is possible to implement more expressive transformations as long as they remain invertible and computationally efficient. This is the case of the so-called rational-quadratic neural spline flows~\cite{durkan2019neural}. 

The spline bijectors are made of $K$ bins, where in each bin one defines a monotonically increasing rational-quadratic function. The~binning is defined on an interval $\mathbb{B}=[-B,B]$, outside of which the function is set to the identity transformation. The~bins are defined by a set of $K+1$ coordinates $\lbrace (x_{i}^{(k)},y_{i}^{(k)})\rbrace^{K}_{k=0}$, called \textit{knots}, strictly monotonically increasing between $\lbrace (x_{i}^{(0)},y_{i}^{(0)})=(-B,-B)$ and $\lbrace (x_{i}^{(K)},y_{i}^{(K)})=(B,B)$. We use the bracket index notation to denote knots' coordinates, which are defined for each dimension of the vectors $x_{i}$ and $y_{i}$. It is possible to construct a rational-quadratic spline bijector with the desired properties with the following procedure~\cite{10.1093/imanum/2.2.123}.

Let us define the quantities

\be
\bry{ll}
h_{i}^{(k)}=x_{i}^{(k+1)}-x_{i}^{(k)}\,,\vspace{2mm}\\
\Delta_{i}^{(k)}=(y_{i}^{(k+1)}-y_{i}^{(k)})/h_{i}^{(k)}\,.
\ery
\ee

Obviously, $\Delta_{i}^{(k)}$ represents the variation of $y_{i}$ with respect to the variation of $x_{i}$ within the $k$-th bin. Moreover, since we assumed strictly monotonically increasing coordinates, $\Delta_{i}^{(k)}$ is always positive or zero. We are interested in defining a bijector $\mathbf{g}(x_{i})$ and mapping the $\mathbb{B}$ interval to itself, such that $\mathbf{g}(x_{i}^{(k)})=y_{i}^{(k)}$, and~with derivatives $d_{i}^{(k)}=d y_{i}^{(k)}/d x_{i}^{(k)}$ satisfying the following conditions:
\be
\bry{ll}
d_{i}^{(k)}=d_{i}^{(k+1)}=0\,\, & {\rm for}\, \Delta_{i}^{(k)}=0\,,\vspace{2mm}\\
d_{i}^{(k)},d_{i}^{(k+1)}>0\,\, & {\rm for}\, \Delta_{i}^{(k)}> 0\,
\ery
\ee
Such condition is necessary and~also sufficient, in~the case of a rational quadratic function, to~ensure monotonicity~\cite{10.1093/imanum/2.2.123}. Moreover, for~the boundary knots, we set $d_{i}^{(0)}=d_{i}^{(K)}=1$ to match the linear behavior outside the $\mathbb{B}$ interval.

For $x_{i}\in [x_{i}^{(k)},x_{i}^{(k+1)}]$, we define

\be
\theta_{i}=(x_{i}-x_{i}^{(k)})/h_{i}^{(k)}\,
\ee
such that $\theta_{i}\in[0,1]$. Then, for~$x_{i}$ in each of the intervals $ [x_{i}^{(k)},x_{i}^{(k+1)}]$ with $k=0,\ldots,K-1$, we define
\be\label{eq:RQS}
y_{i}=P_{i}^{(k)}(\theta_{i})/Q_{i}^{(k)}(\theta_{i})\vspace{1mm}\,
\ee
with the functions $P$ and $Q$ defined by
\small\be
\bry{lll}
P_{i}^{(k)}(\theta_{i})&=&\Delta_{i}^{(k)}y_{i}^{(k+1)}\theta_{i}^{2}+\Delta_{i}^{(k)}y_{i}^{(k)}(1-\theta_{i})^{2}\vspace{2mm}\\
&&+(y_{i}^{(k)}d_{i}^{(k+1)}+y_{i}^{(k+1)}d_{i}^{(k)})\theta_{i}(1-\theta_{i})\,,\vspace{2mm}\\
Q_{i}^{(k)}(\theta_{i})&=&\Delta_{i}^{(k)}+(d_{i}^{(k+1)}+d_{i}^{(k)}-2\Delta_{i}^{(k)})\theta_{i}(1-\theta_{i})\,
\ery
\ee\normalsize

The ratio in Eq.~\eqref{eq:RQS} can then be written in the simplified form
\small\be\label{eq:RQS1}
y_{i}=y_{i}^{(k)}+\frac{(y_{i}^{(k+1)}-y_{i}^{(k)})(\Delta_{i}^{(k)}\theta_{i}^{2}+d_{i}^{(k)}\theta_{i}(1-\theta_{i}))}{\Delta_{i}^{(k)}+(d_{i}^{(k+1)}+d_{i}^{(k)}-2\Delta_{i}^{(k)})\theta_{i}(1-\theta_{i})}\,
\ee\normalsize

The Jacobian $\textrm{J}_{\mathbf{g}}=\partial y_{i}/\partial x_{j}$ is then diagonal, with~entries given by
\small\be\label{eq:rq-derivative}
 \frac{(\Delta_{i}^{(k)})^{2} (d_{i}^{(k+1)} \theta_{i}^{2} + 2 \Delta_{i}^{(k)} \theta_{i} (1 - \theta_{i}) + d_{i}^{(k)} (1 - \theta_{i})^{2} )}{(\Delta_{i}^{(k)} +(d_{i}^{(k+1)} + d_{i}^{(k)} - 2 \Delta_{i}^{(k)}) \theta_{i} (1 - \theta_{i}))^{2}}\,
\ee\normalsize
for $i=1,\ldots,D$. The~inverse of the transformation \eqref{eq:RQS} can also be easily computed by solving the quadratic Eq.s \eqref{eq:RQS} with respect to $x_{i}$.

In practice, $B$ and $K$ are hyperparameters, while $\lbrace (x_{i}^{(k)},y_{i}^{(k)})\rbrace^{K}_{k=0}$ and $\lbrace d_{i}^{(k)}\rbrace_{k=1}^{K-1}$ are $2(K+1)$ plus $K-1$ parameters, modeled by an NN, which determine the shape of the spline function. The~different implementations of the RQS bijector, in~the context of coupling and autoregressive flows, are determined by the way in which such parameters are computed. We briefly describe them in turn in the following two~subsections.

\subsection{The~C-RQS}
In the coupling flow case (C-RQS), one performs the usual partitioning of the $D$ dimensions in the two sets composed of the first $d$ and last $D-d$ dimensions. The~first $d$ dimensions are then kept unchanged $y_{i}=x_{i}$ for $i=1,\ldots,d$, while the parameters describing the RQS transformations of the other $D-d$ dimensions are determined from the inputs of the first $d$ dimensions, denoted by $x_{1:d}$. Schematically, we could write

\be
\bry{lll}
&x_{i}^{(k)}&=x_{i}^{(k)}(x_{1:d})\,,\vspace{2mm}\\
&y_{i}^{(k)}&=y_{i}^{(k)}(x_{1:d})\,,\vspace{2mm}\\
&d_{i}^{(k)}&=d_{i}^{(k)}(x_{1:d})\,
\ery
\ee
for $i=d+1,\ldots,D$.

A schematic description of our implementation of the C-RQS is given in \mbox{Appendix \ref{app:CRQS}}.

\subsection{The~A-RQS}
The RQS version of the MAF, which we call A-RQS, is instead obtained by leaving unchanged the first dimension $y_{1}=x_{1}$ and determining the parameters of the transformation of the $i$-th dimension from the output of all preceding dimensions, denoted by $y_{1:i-1}$. Schematically, this is given by

\be
\bry{lll}
&x_{i}^{(k)}&=x_{i}^{(k)}(y_{1:i-1})\,,\vspace{2mm}\\
&y_{i}^{(k)}&=y_{i}^{(k)}(y_{1:i-1})\,,\vspace{2mm}\\
&d_{i}^{(k)}&=d_{i}^{(k)}(y_{1:i-1})\,
\ery
\ee
for $i=2,\ldots,D$. 

\section{Non-Parametric Quality~Metrics}\label{sec:metrics}
We assessed the performance of our trained models using three distinct metrics: the dimension-averaged 1D Kolmogorov--Smirnov (KS) two-sample test statistic $\overline{D}_{y,z}$, the~sliced Wasserstein distance (SWD) $\overline{W}_{y,z}$, and~the Frobenius norm (FN) of the difference between the correlation matrices of two samples $\|C\|_{F} $. With~a slight abuse of nomenclature, we refer to these three different distance measures with~vanishing optimal value simply as KS, SWD, and~FN, respectively. Each of these metrics served as a separate test statistic in a two-sample test, where the null hypothesis assumed that both samples originate from the same target distribution. For~each metric, we established its distribution under the null hypothesis by drawing both samples from the target distribution. We then compared this distribution with the test-statistic calculated from a two-sample test between samples drawn from the target and NF distributions to assign each model a $p$-value for rejecting the null~hypothesis.

To quantify the uncertainty on the test-statistics computed for the test vs NF-generated samples, we performed the tests 10 times using differently seeded target- and NF-generated samples. We calculated \emph{}p-values based on the mean test statistic and its $\pm 1$ standard~deviation.

For model comparison and~to assess the uncertainty on the training procedure, we trained 10 instances of each model configuration, defined by a set of hyperparameter values. Rather than selecting the single best-performing instance to represent the best architecture, we averaged the performances across these instances and identified the architecture with the best average performance. After~selecting this top-performing model, which we called the ``best'' model, we reported both its average and peak~performances.

When computing the test-statistics distributions under the null hypothesis and~evaluating each model's \emph{p}-values, we found that the discriminative power of the KS metric was larger than that of the SWD and FN ones. For~this reason, we used the result of the KS-statistic to determine the best model and then showed results also for the other two statistics. Nevertheless, even though the best model could vary depending on the metric used, no qualitative difference in the conclusions would arise from choosing FN or SWD as the ranking metric; i.e.,~results were not identical but were consistent among the three~metrics.

It is important to stress that despite our insistence on using non-parametric quality metrics, we actually know the target density, and~we used this information for bootstrapping uncertainties and computing \emph{p}-values. In~real-world examples, the target density is generally not known, and depending on the number of available samples, our procedure for evaluation needs to be adapted or may end up being unusable. Nevertheless, this well-defined statistical approach is crucial for us since we aim to draw rather general conclusions, which strongly depend on the ability to estimate the uncertainties and should rely on robust statistical~inference.

In the following, we briefly introduce the three aforementioned metrics. To~do so, we employ the following notation: we indicate with $N$ the number of $D$-dimensional points in each sample and use capital indices $I,J$ to run over $N$ and lowercase indices $i,j$ to \mbox{run over $D$}.\footnote{We warn the reader not to confuse the dimensionality$D$ with the KS test-statistic $D_{x,y}$.} We also use Greek letters indices $\alpha,\beta$ to run over slices (random directions).

\begin{figure*}[t!]
\includegraphics[scale=.45]{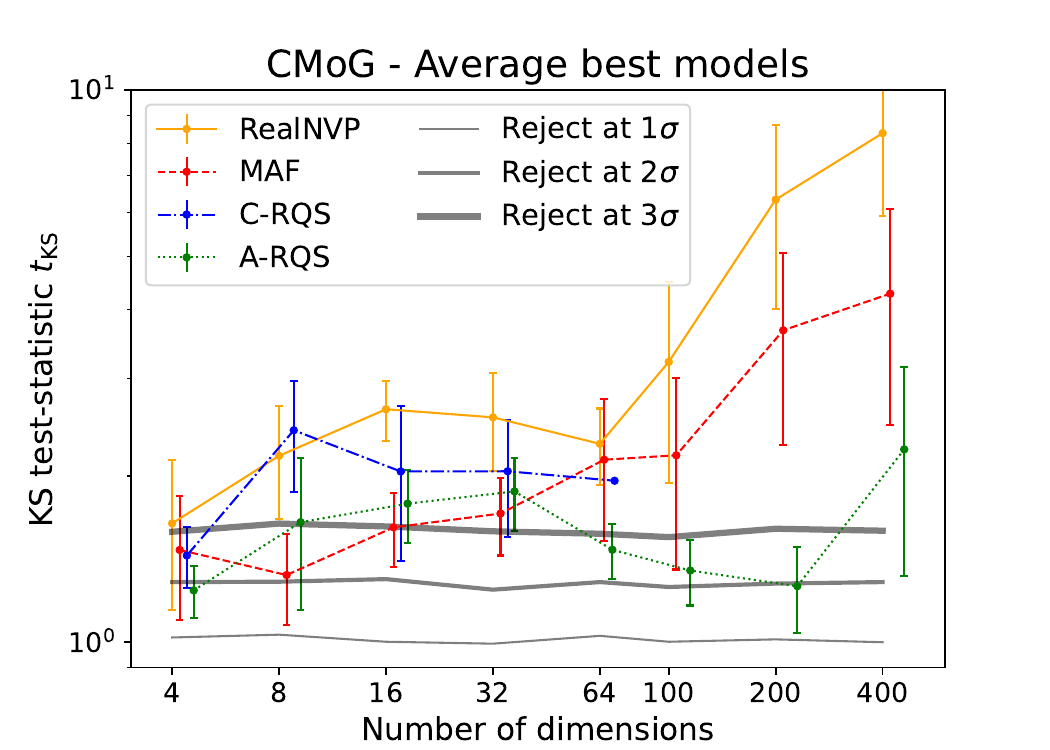}
\includegraphics[scale=.45]{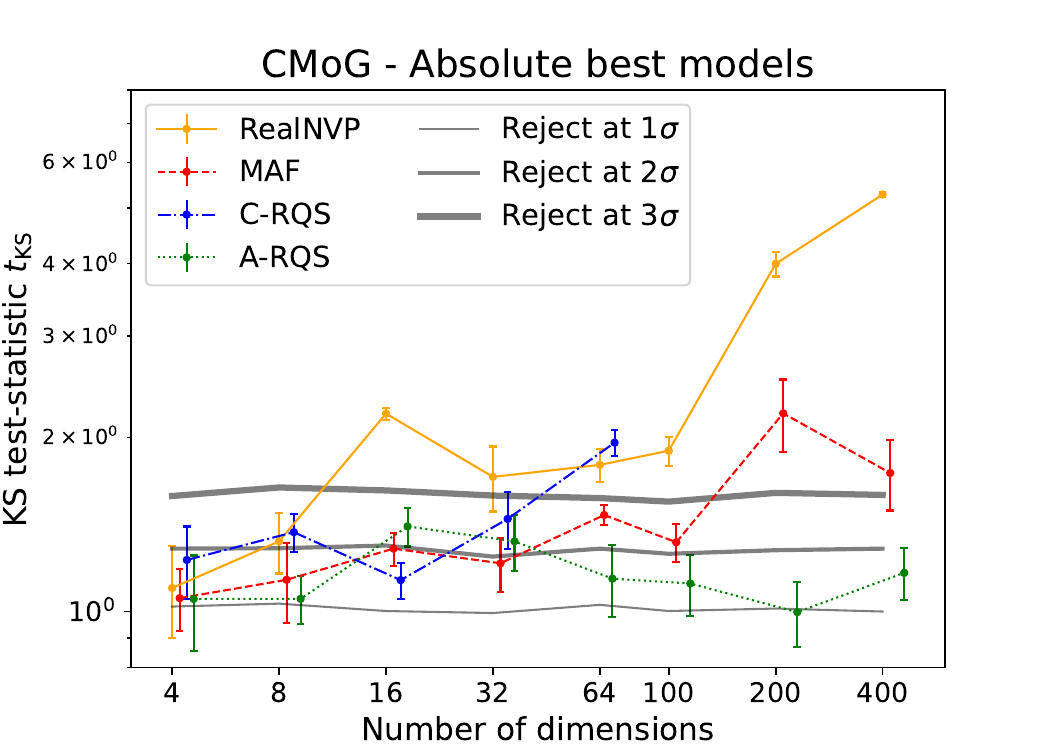}
\includegraphics[scale=.45]{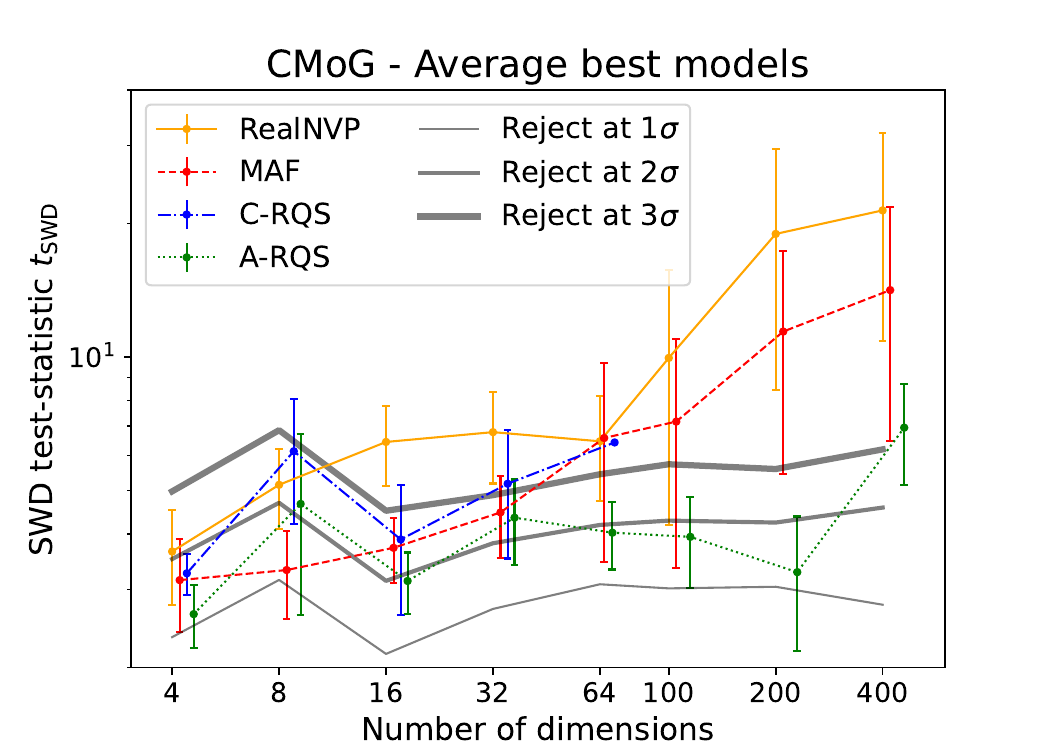}
\includegraphics[scale=.45]{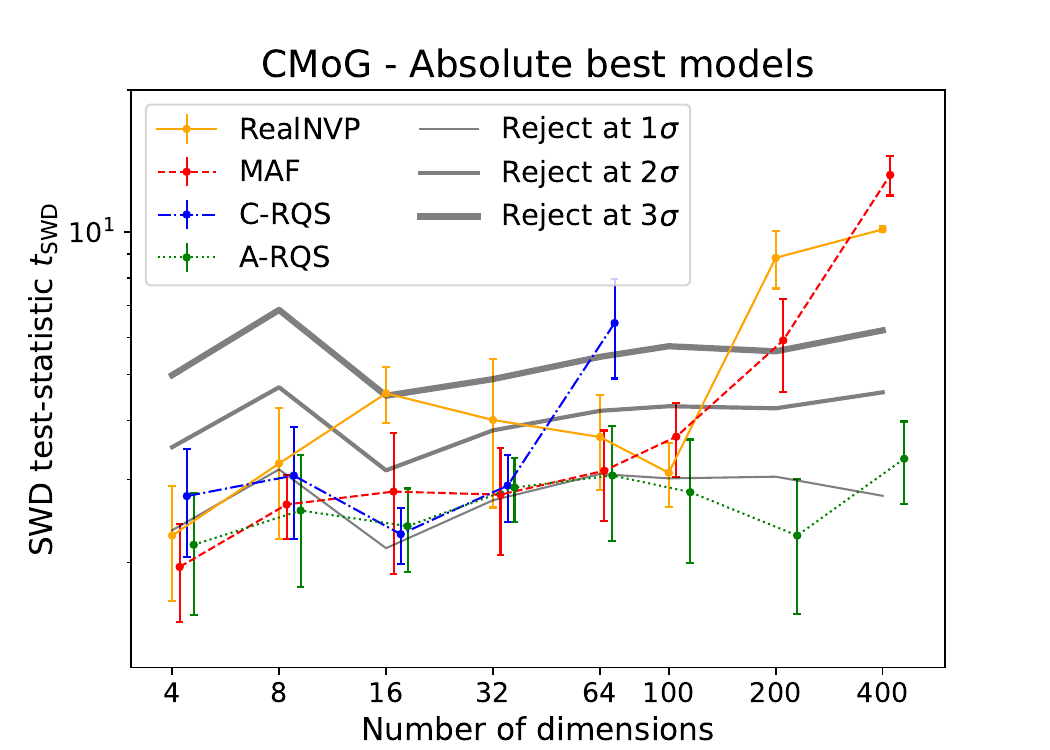}
\includegraphics[scale=.45]{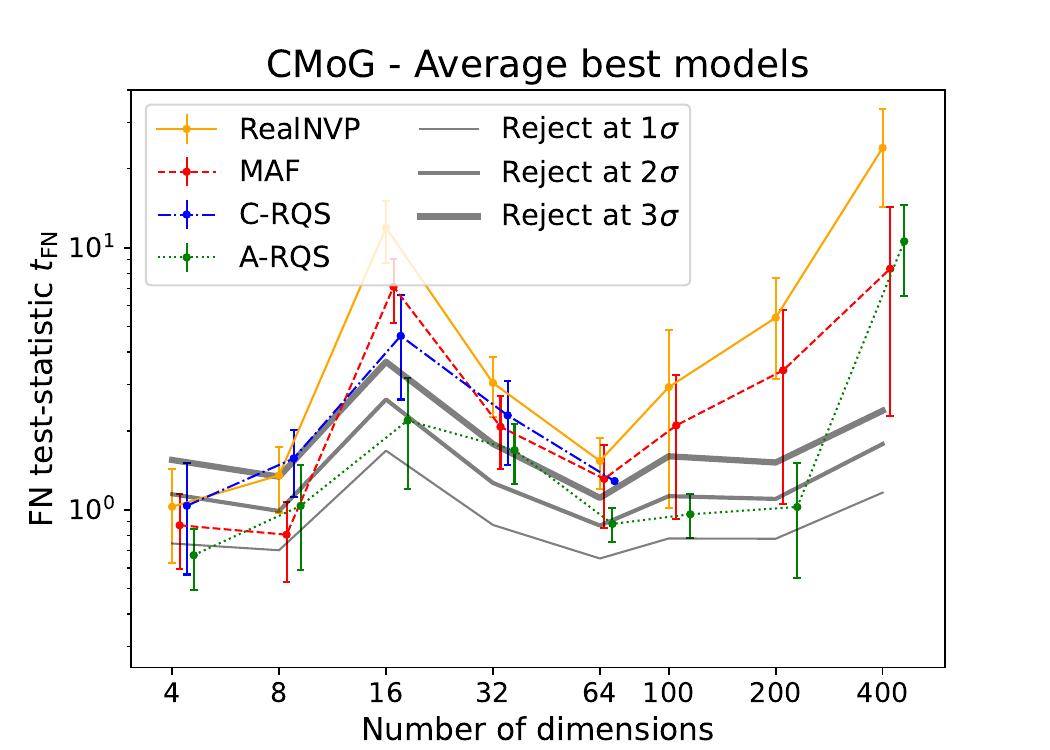}
\includegraphics[scale=.45]{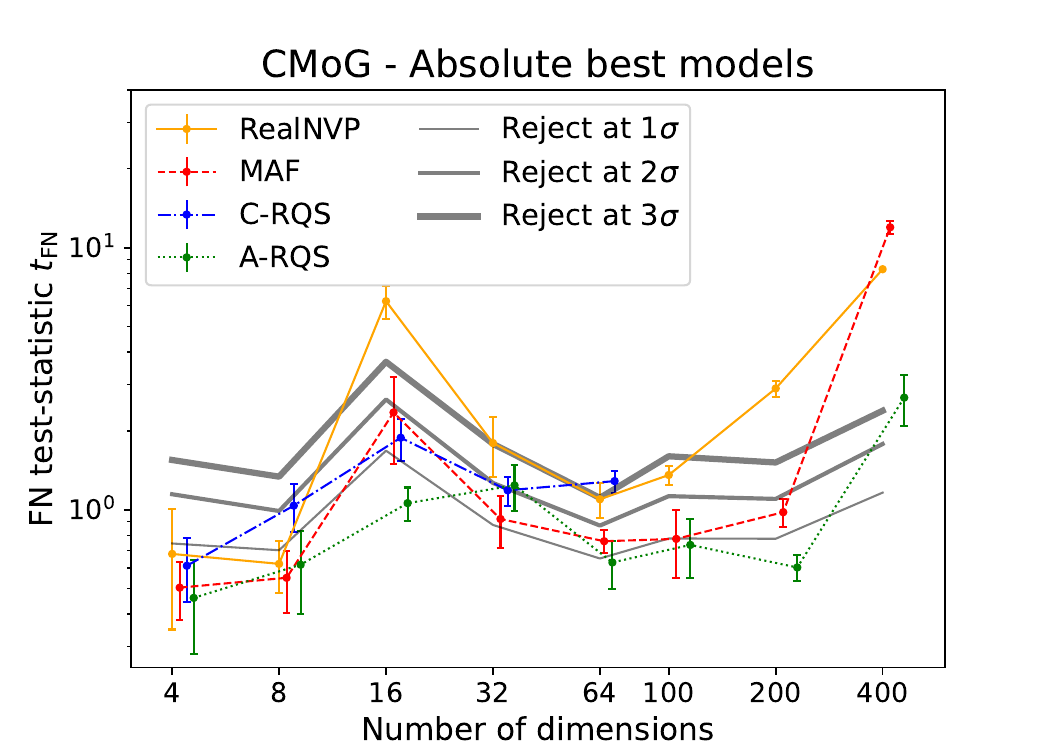}
\caption{Performance comparison between the average (\textbf{left panel}) and absolute (\textbf{right panel}) best models obtained with RealNVP, MAF, C-RQS, and A-RQS architectures when learning the CMoG distributions. The figures show the value of the test statistic with its uncertainty, computed as explained in the text. The KS, SWD, and FN test statistics, as defined in Section \ref{sec:metrics}, are shown in the upper, middle, and lower panel, respectively. The gray lines with different widths represent, from thinner to ticker, the $1,2,3\sigma$ thresholds for the test statistics, as obtained from the test-statistic distributions under the null hypothesis, evaluated with $10^{4}$ pseudo-experiments.}
\label{fig:cmog_metrics}
\end{figure*}

\begin{itemize}
\item \textbf{Kolmogorov--Smirnov test}
\\
The KS  test is a statistical test used to determine whether or not two 1D samples are drawn from the same \textit{unknown} PDF. The~null hypothesis is that both samples come from the same PDF. The~KS test statistic is given by
\begin{equation}
D_{y,z}=\mathbf{sup}_{x}\mid F_{y}(x)-F_{z}(x)\mid\,,
\end{equation}
where $F_{y,z}(x)$ are the empirical distributions of each of the samples $\{y_{I}\}$ and $\{z_{I}\}$, and~$\mathbf{sup}$ is the supremum function. For~characterizing the performances of our results, we computed the KS test-statistic for each of the 1D marginal distributions along the $D$ dimensions and took the average as follows:

\vspace{-12pt}
\begin{equation}
\overline{D}_{y,z} = \frac{1}{D}\sum_{i=1}^{D} D_{y,z}^{i}\,.\vspace{1mm}
\end{equation}

The actual test statistic that we consider in this paper is the scaled version of $\overline{D}_{y,z}$, given by

\vspace{-12pt}
\begin{equation}
t_{\rm{KS}} = \sqrt{\frac{N}{2}}\overline{D}_{y,z}\
\end{equation}

The $\sqrt{N/2}$ factor comes from the known $\sqrt{m\cdot n/(m+n)}$ factor in the scaled KS statistic with different-sized samples of~sizes $m$ and $n$, respectively. 

Notice that even though the test statistic

\begin{equation}
\sqrt{\frac{m\cdot n}{m+n}}D_{y,z}
\end{equation}
is asymptotically distributed according to the Kolmogorov distribution~\cite{KolmogorovSmirnov1933SullaDE,Smirnov1939,Kolmogoroff1941ConfidenceLF,Smirnov1948TableFE,ksscipy}, the~same is not true for our $t_{\rm{KS}}$ statistic due to correlations among dimensions. Nevertheless, our results seem to suggest that the asymptotic distribution of $t_{\rm{KS}}$ for large $D$ (that means when the average is taken over many dimensions) has a reasonably universal behavior, translating into almost constant rejection lines (solid gray lines with different thicknesses) in the upper panels of Figure~\ref{fig:cmog_metrics}.

\item \textbf{Sliced Wasserstein Distance} \\
The SWD~\cite{Rabin2011WassersteinBA,Bonneel2014} is a distance measure for comparing two multidimensional distributions based on the 1D Wasserstein distance~\cite{kantorovich1942translocation, wasserstein1969markov}. The~latter distance between two univariate distributions is given as a function of their respective empirical distributions as follows

\begin{equation}
W_{y,z}=\int_{\mathbb{R}}dx \mid F_{y}(x)-F_{z}(x)\mid
\end{equation}
Intuitively, the difference between the WD and the KS test statistic is that the latter considers the maximum distance, while the former is based on the integrated~distance.

Our implementation of the SWD is defined as follows. For~each model and each dimensionality $D$, drew $2D$ random directions $\hat{v}^{i}_{\alpha}$, with~$i=1,\ldots D$ and $\alpha=1,\ldots,2D$, uniformly distributed on the surface of the unit $N$-sphere\footnote{This can be done by normalizing the $D$-dimensional vector obtained by sampling each components from an independent standard normal distribution~\cite{10.1145/377939.377946}.} Given two $D$-dimensional samples $\{\mathbf{y}_{I}\}=\{y_{I}^{i}\}$ and $\{\mathbf{z}_{I}\}=\{z_{I}^{i}\}$, we considered the $2D$ projections

\be
\{y_{I}^{\alpha}\} = \left\{\sum_{i=1}^{D}y_{I}^{i}\hat{v}_{i}^{\alpha}\right\} \,,\qquad \{z_{I}^{\alpha}\} = \left\{\sum_{i=1}^{D}z_{I}^{i} \hat{v}_{i}^{\alpha}\right\}\,
\ee
and computed the corresponding $2D$ Wasserstein distances as follows:

\begin{equation}
W_{y,z}^{\alpha}=\int_{\mathbb{R}}dx \mid F_{y^{\alpha}}(x)-F_{z^{\alpha}}(x)\mid
\end{equation}

was

\be
\overline{W}_{y,z}=\frac{1}{2D}\sum_{\alpha=1}^{2D} W_{y,z}^{\alpha}\,.
\ee
In analogy with the scaled KS test statistic, we defined the scaled SWD test statistic as follows:

\begin{equation}
t_{\rm{SWD}} = \sqrt{\frac{N}{2}}\overline{W}_{y,z}\,
\end{equation}

 \vspace{2mm}

\item \textbf{Frobenius norm} \\
The FN of a matrix $M$ is given by

\begin{equation}
\| M \|_{F}= \sqrt{\sum_{i,j} \mid m_{ij}\mid ^2},
\end{equation}
where $m_{ij}$ are the elements of $M$. By~defining $C=C_{y}-C_{z}$, where $C_{y},C_{z}$ are the two $N\times N$ correlation matrices of the samples $\{\mathbf{y}_{I}\}=\{y_{I}^{i}\}$ and $\{\mathbf{z}_{I}\}=\{z_{I}^{i}\}$, its FN, given by
\begin{equation}
\| C \|_{F}= \sqrt{\sum_{i,j} \mid c_{y,ij}-c_{z,ij}\mid^2}
\end{equation}
is a distance measure between the two correlation matrices. 
In analogy with the previously defined test statistics, we defined the scaled FN test statistic as follows:
\begin{equation}
t_{\rm{FN}} = \sqrt{\frac{N}{2}}\frac{\| C \|_{F}}{D}\,,
\end{equation}
where we also divided by the number of dimensions $D$ to remove the approximately linear dependence of the FN distance on the dimensionality of the~samples.
\end{itemize}

\section{Testing the Normalizing~Flows}\label{Sec:Tests}
We tested the four architectures discussed above on CMoG distributions defined as a mixture of $n=3$ components and $D= 4,8,16,32,64,100,200,400$ dimensional multivariate Gaussian distributions with diagonal covariance matrices, parametrized by means randomly generated in the $[0,10]$ interval and standard deviations randomly generated in the $[0,1]$ interval\footnote{The values for the means and standard deviations were chosen so that the different components could generally be resolved.} The~components were mixed according to an $n$ dimensional categorical distribution (with random probabilities). This meant that a different probability was assigned to each component, while different dimensions of the same component multivariate Gaussian were assigned the same probability. The resulting multivariate distributions had random order-one off-diagonal elements in the covariance matrix and multi-modal 1D marginal distributions (see, for~illustration, \mbox{Figures~\ref{fig:corre_matrix_plot} and \ref{fig:corner_corr_100}}).

In our analysis, we considered a training set of $10^5$ points, a~validation set of $3 \times 10^4$ points, and~a test set equal in size to the training set, with~$10^5$ points. It is important to note that the chosen size of the test set corresponds to the most stringent condition for evaluating the NF models. This is because the NF cannot be expected to approximate the real target distribution more accurately than the uncertainty determined by the size of the training~sample.

In practical terms, the most effective NF would be indistinguishable from the target distribution when tested on a sample size equivalent to the training set. In~our analysis, we found that models tested with $10^5$ samples often led to rejection at the $2\sigma/3\sigma$ level, at~least with the most powerful KS test. However, this should not be viewed as a poor outcome. Rather, it suggests that one needs to utilize a test set as large as the training set to efficiently discern the NF from the true model, while smaller samples are effectively indistinguishable from those generated with the target~distribution.

An alternative approach, which we did not adopt due to computational constraints, involves calculating the sample size required to reject the null hypothesis at a given confidence level. This approach offers a different but equally valid perspective, potentially useful for various applications. Nevertheless, our approach was efficient for demonstrating that NFs can perform exceptionally well on high-dimensional datasets and for comparing, among~each other, the~performances of different NF~architectures.

For each of the four different algorithms described above and for each value of $N$, we performed a small scan over some of the free hyperparameters. Details on the choice of the hyperparameters are reported in Appendix \ref{app:mog_hyperparam}. 
All models were trained on Nvidia A40~GPUs. 

The performances of the best NF architectures are reported in Figure~\ref{fig:cmog_metrics} and \linebreak  Tables~\ref{tab:average_best} and  \ref{tab:absolute_best}.

Figure \ref{fig:cmog_metrics} shows the values of the three test statistics (vertical panels) for the average (left panels) and absolute (right panels) best models obtained with the four different architectures. The~three gray lines with different thicknesses represent the values of the test statistics corresponding to $1\sigma$,  $2\sigma$, and~$3\sigma$ rejection ($p$-values of $0.68$, $0.95$, and~$0.99$, respectively) of the null hypothesis that the two samples (test and NF-generated) are drawn from the same PDF. These rejection lines were obtained through $10^{4}$ pseudo-experiments. The~curve for the best C-RQS models stops at $64$D since the training becomes unstable and the model does not converge. The~situation could likely be improved by adding regularization and by fine-tuning the hyperparameters. However, to~allow for a fair comparison with the other architectures, where regularization and fine-tuning are not necessary for a reasonable convergence, we avoided pushing C-RQS beyond $64$D. Also notice that the uncertainty shown in the point at $64$D for the C-RQS is artificially very small since only a small fraction of the differently seeded runs converged. This uncertainty should therefore be considered~unreliable. 

All plots in Figure~\ref{fig:cmog_metrics} include uncertainties. As~already mentioned, the~best model was chosen as the one with best architecture on average, and~therefore, over~$10$ different trainings were performed with differently seeded training samples.
For the selection of the best model, the~left plots show the performances averaged over the $10$ trainings, with~error bands representing the corresponding standard deviations, while the right plots show the performances of the absolute best instance among the $10$ trained replicas, with the~error band representing the standard deviation over the $10$ replicas generated for testing (test and NF-generated samples). In~other words, we can say that the uncertainties shown in the left plots are the standard deviations due to repeated training, while the uncertainties shown in the right plots are the standard deviations due to repeated generation/evaluation (testing).

Figure~\ref{fig:cmog_metrics} clearly highlights the distinct characteristics that establish the A-RQS as the top-performing algorithm:
\bit
    \item Its performances are almost independent of the data dimensionality;
    \item The average best model is generally not rejected at $3\sigma$ level when evaluated with a number of points equal to the number of training points;
    \item The absolute best model is generally not rejected at $2\sigma$ level when evaluated with a number of points equal to the number of training points;
    \item The uncertainties due to differently seeded training and testing are generally comparable, while for all other models, the~uncertainty from training is generally much larger than the one from evaluation.
\eit

All values shown in Figure~\ref{fig:cmog_metrics} are reported in Tables~\ref{tab:average_best} and  \ref{tab:absolute_best} for~the average and absolute best models, respectively. In~the tables, we also show the total number of trainable parameters, the~average number of epochs, training time, and~prediction time. It is interesting to look at the training and prediction times. Indeed, while for the coupling flows, even though training time is much larger than prediction time, both times grow with a similar rate; for~the autoregressive flows, the~prediction time grows faster than does the training time, which is almost constant. This is because, as~we have already mentioned, the~MAF is a ``direct flow'', very fast for density estimation and~therefore for training (single pass through the flow) but~slower for generation and~therefore for testing ($N$ passes through the flow, with~$N$ the dimensionality of the target distribution). Still, testing was reasonably fast, considering that each test actually consisted of $10$ tests with three metrics and $10^{5}$ points per sample. All trainings/tests took less than a few hours (sometimes, especially in small dimensionality, a~few minutes), which means that all models, expect~the C-RQS in large dimensionalities, are fairly fast both in training and inference.\footnote{Notice that even though the~training/testing times do not go beyond a few hours, we trained and tested $10$ replicas of four architectures in eight different dimensionalities (apart from C-RQS) and with a few different values of the hyperparameters for~a total of about $1360$ runs (see Table~\ref{tab:hyperparam}). This took several months of GPU time, showing how resource demanding is to reliably estimate uncertainties of ML models, even in relatively simple cases.} Another interesting number in the tables is the total number of trainable parameters. Such number makes clear how the autoregressive architectures are more expressive than are the simple coupling ones, giving better results with relatively fewer parameters. It is also clear from the table that the improvement stepping from a simple linear affine bijector to a rational quadratic spline based on the same architecture is much larger for autoregressive architectures and~less evident for the simplest coupling ones. The~large number of parameters needed to obtain reasonable results from C-RQS may be the origin of its training instability at large~dimensionality.

\section{Conclusions and~Outlook}\label{ref:conclusion}
Normalizing flows have shown many potential applications in a wide variety of research fields including HEP, both in their normalizing and generative directions. However, to~ensure a standardized usage and to match the required precision, their application to high-dimensional datasets need to be properly evaluated. This paper makes a step forward in this direction by quantifying the ability of coupling and autoregressive flow architectures to model distributions of increasing complexity and~dimensionality. 

Our strategy consisted in performing statistically robust tests utilizing different figures of merits and~including estimates of the variances induced both by the training and the evaluation~procedures. 

We focused on the most widely used NF architectures in HEP, the~coupling (RealNVP and C-RQS) and the autoregressive flows (MAF and A-RQS), and~compared them against generic multimodal distributions of increasing~dimensionality. 

As the main highlight, we found that the A-RQS is highly capable of precisely learning all the high-dimensional complicated distributions it was tested against, always within a few hours of training on a Tesla A40 GPU and with limited training data. Moreover, the~A-RQS architecture, showed great generalization capabilities, obtaining almost constant results over a very wide range of dimensionalities, ranging from $4$ to $400$.

As for the other tested architectures, our results show that reasonably good results can be obtained with all of them but the C-RQS, which ended up being the least capable in generalizing to large dimensionality, with~unstable and longer trainings, especially in high~dimensionality. 

Our analysis was performed implementing all architectures in \textsc{TensorFlow2} with \textsc{TensorFlow Probability} using \textsc{Python}. The~code is available in Ref.~\cite{github1}, while a general-purpose user-friendly framework for NFs in TensorFlow2 named \textsc{NF4HEP} is under development and can be found in Ref.~\cite{github2}. Finally, a~code for statistical inference and two-sample tests, implementing the metrics considered in this paper (and others) in \textsc{TensorFlow2}, is available in Ref.~\cite{github3}.

We stress that the intention of this study was to secure generic assessments of how NFs perform in high dimensions. For~this reason, the target distributions were chosen independently of any particular experimentally driven physics dataset. An~example of application to a physics dataset, in~the direction of building an unsupervised DNNLikelihood~\cite{Coccaro:2019lgs}, is presented in Ref.~\cite{Reyes-Gonzalez:2023oei}. Nonetheless, these studies represent the firsts of a series to come. Let us briefly mention, in~turn, the~research directions we aim to follow starting from the present~paper.
\begin{itemize}
\item Development of reliable multivariate quality metrics, including approaches based on machine learning~\cite{Friedman:2003id,Kansal:2022spb}. 
We note the importance of performing statistically meaningful tests on generative models, ideally including uncertainty estimation. A~thorough study of different quality metrics against high dimensional data is on its way. Moreover, new results~\cite{DAgnolo:2019vbw,Chakravarti:2021svb,Letizia:2022xbe} suggest that classifier-based two-sample tests have the potential to match the needs of the HEP community when paired with a careful statistical analysis. These tests can leverage different ML models to provide high flexibility and sensitivity together with short training times, especially when based on kernel methods~\cite{Letizia:2022xbe}. On~the other hand, further studies are needed to investigate their efficiency and scalability to high dimensions.
\item Study of the dependence of the NF performances on the size of the training sample~\cite{DelDebbio:2021mts}. In~the present paper, we always kept the number of training points to $10^{5}$. It is clear that such a number is fairly large in small dimensionality, such as $N=4$ dimensions, and~undersized for large dimensionality, such as $N\geq 100$. It is important to study the performances of the considered NF architectures in the case of scarce or very abundant data and to assess the dependence of the final precision on the number of training samples. This can also be related to developing techniques to infer the uncertainty of the NF models.
\item Studies on how NFs can be used for statistical augmentation. For~instance, NFs can be used for normalizing direction to build effective priors and proposals to enhance (in terms of speed and convergence time) known sampling techniques, such as Markov chain Monte Carlo, whose statistical properties are well established.
\item The ability to preserve and distribute pretrained NF-based models. A final issue that needs to be addressed to ensure a widespread use of NFs in HEP is the ability to preserve and distribute pretrained NF-based models. This is, for~the time being, not an easy and standard task, and support from the relevant software developers in the community is crucial to achieving this goal.
\end{itemize}

\subsection*{Acknowledgments}
We thank Luca Silvestrini for pushing us towards the study of NFs and for fruitful discussions. We are thankful to the developers of \textsc{TensorFlow2}, which made possible to perform this analysis. We thank the IT team of INFN Genova, and in particular Mirko Corosu, for useful support. We are thankful to OpenAI for the development of ChatGPT and to ChatGPT itself for useful discussions. H.R.G. acknowledges the hospitality of Sabine Kraml at LPSC Grenoble and the discussions on Normalizing Flows held there with the rest of the SModelS Collaboration. This work was partially supported by the Italian PRIN grant 20172LNEEZ. M.L. acknowledges the financial support of the European Research Council (grant SLING 819789). H.R.G. is supported by the Deutsche Forschungsgemeinschaft (DFG, German Research Foundation) under grant 396021762 – TRR 257: Particle Physics Phenomenology after the Higgs Discovery

\appendix
\section{Implementation of NF Architectures}\label{app:Algo}

\subsection{The~RealNVP}\label{app:RealNVP}
We are given a collection of vectors $\{y_{i}^{I}\}$ with $i=1,\ldots,D$ representing the dimensionality and $I=1,\ldots,N$ the number of samples representing the unknown PDF $p_{Y}$. For~all samples $\mathbf{y}^{I}$, we consider the half partitioning given by the two sets $\mathbf{\hat{y}}^{I}=\{y_{1}^{I},\ldots,y_{D/2}^{I}\}$ and $\mathbf{\tilde{y}}_{I}=\{y_{D/2+1}^{I},\ldots,y_{D}^{I}\}$ (For simplicity, we assume $D$ is even and therefore $D/2$ integer. In~case $D=2d+1$ is odd the ``half-partitioning'' could be equally done by taking the first $d+1$ and the last $d$ dimensions, or~vice versa. This does not affect our implementation). We then use the $\mathbf{\hat{y}}^{I}$ samples as inputs to train a fully connected MLP (a dense DNN) giving as output the vectors of $t_{i}$ and $s_{i}$, with~$i=1,\ldots,D/2$ in Eq.~\eqref{eq:realnvp}. These output vectors are provided by the DNN through two output layers, which are dense layers with linear and $\tanh$ activation functions for $t_{i}$ and $s_{i}$, respectively, and~are used to implement the transformation in Eq.~\eqref{eq:realnvpinverse}, which outputs the (inversely) transformed samples. Moreover, in~order to transform all dimensions and to increase the expressivity of the model, we use a series of such RealNVP bijectors, feeding the output of each bijector as input for the next one and~inverting the role of the two partitions at each step. After~the full transformation is performed, one obtains the final $\{x_{i}^{I}\}$ with $i=1,\ldots,D$ vectors and the transformation Jacobian (the product of the inverse of Eq.~\eqref{eq:realnvpjacobian} for each bijector). With~these ingredients, and~assuming a normal base distribution $p_{X}$, one can compute the negative of the log-likelihood in Eq.~\eqref{loss}, which is used as loss function for the DNN~optimization.

As is clear from the implementation, the~RealNVP NF, i.e.,~the series of RealNVP bijectors, is trained in the normalizing direction, taking data samples as inputs. Nevertheless, since the $s_{i}$ and $t_{i}$ vectors only depend, at~each step, on~untransformed dimensions, once the DNN is trained, they can be used both to compute the density by~using Eq.~\eqref{eq:realnvpinverse} and to generate new samples, with~equal efficiency, by~using Eq.~\eqref{eq:realnvp}. This shows that the RealNVP is equally efficient in both the normalizing and generative~directions.

\subsection{The~MAF}\label{app:MAF}
As in the case of the RealNVP,  for the MAF, the~forward direction represents the normalizing direction. In~this case, the vectors $s_{i}$ and $t_{i}$ of~dimension $D-1$ describing the affine bijector in Eq.~\eqref{eq:MAF1} are parametrized by an autoregressive DNN with $D$ inputs and $2(D-1)$ outputs, implemented through the MADE~\cite{MADE2015} masking procedure according to the \textsc{TensorFlow Probability} implementation (see Ref.~\cite{TFPMAF}). The~procedure is based on binary mask matrices that define which connections (weights) are kept and which are dropped to ensure the autoregressive property. (The binary mask matrices are simple transition matrices between pairs of layers of dimension $(K',K)$, with $K'$ being the number of nodes in the forward layer (closer to the output) and $K$ being the number of nodes in the backward layer (closer to the input). Obviously $K=D$ is for the input layer, and $K'=2(D-1)$ is for the output  layer.) Mask matrices are determined from numbers (degrees) assigned to all nodes in the DNN: each node in the input layer is numbered sequentially from $1$ to $D$; each node in each hidden layer is assigned a number between $1$ and $D$, possibly with repetition; the first half output nodes (representing $s_{i}$) are numbered sequentially from \mbox{$1$ to $D-1$}, and the same is done for the second half (representing $t_{i}$). Once all degrees are assigned, the~matrix elements of the mask matrices are $1$ if two nodes are connected and $0$ if they are ``masked'', i.e.,~not connected. The~mask matrices are determined by connecting the nodes of each layer with index $k$ with all nodes in the preceding layer that have an index smaller or \mbox{equal than $k$}. As~for the RealNVP, a~series of MAF bijectors is used, by~feeding each with the $\{x_{i}^{I}\}$, with~$i=1,\ldots,D$, according to Eq.~\eqref{eq:MAF1inverse} computed from the previous one. The~last bijector computes the final $\{x_{i}^{I}\}$, with~$i=1,\ldots,D$, according to Eq.~\eqref{eq:MAF1inverse} and the transformation Jacobian (the product of the inverse of Eq.~\eqref{eq:MAF1Jacobian} for each bijector), used to compute and optimize the log-likelihood as defined in Eq.~\eqref{loss}.

The efficiency of the MAF in the normalizing and generative directions is not the same as~in the case of the RealNVP. Indeed, computing the log-likelihood for density estimation requires a single forward pass of $\{y_{i}\}$ through the NF. However, generating samples requires a start from $\{x_{i}\}$, randomly generated from the base distribution. Then, one needs the following procedure to compute the corresponding $\{y_{i}\}$:
\bit
\item Define the first component of the required $y_{i}^{\text{input}}$ as $y_{1}^{\text{output}}=y_{1}^{\text{input}}=x_{1}$, where $y_{i}^{\text{input}}$ is the NF input;
\item Start with a $y_{i}^{\text{input}}=x_{i}$ and pass it through the NF to determine $y_{2}^{\text{output}}$ as a function of $y_{1}^{\text{output}}$; 
\item Update $y_{i}^{\text{input}}$ with $y_{2}^{\text{input}}=y_{2}^{\text{output}}$ and pass through the NF to determine $y_{3}^{\text{output}}$ as a function of $y_{1}^{\text{output}}$ and $y_{2}^{\text{output}}$;
\item Iterate until all the $y_{i}^{\text{output}}$ components are computed.
\eit

It is clear to see that the procedure requires $D$ to pass through the NF to generate a sample, and so the generation in the MAF is D times less efficient than is the density estimation. The~inverse autoregressive flow (IAF) \cite{10.5555/3157382.3157627} is an implementation similar to the MAF that implements generation in the forward direction (obtained by exchanging $x$ and $y$ in Eq.s \eqref{eq:MAF1} and \eqref{eq:MAF1inverse}. In~the case of IAF, computing the log-likelihood (which is needed for training) requires $D$ steps, while generation only requires a single pass through the flow. The~IAF is therefore much slower in training and much faster in generating new~samples.

\begin{table}[t!]
\footnotesize
\begin{tabular}{lllll}
\toprule
\multicolumn{5}{l}{\bf Hyperparameters values} \\
\toprule
Hyperpar.		&  MAF 		&  RealNVP 		&  A-RQS 	&  C-RQS 		\\
\midrule
number of		& $5,10$		& $5,10$ 		& $2$ 	& $5,10$		\\
bijectors  		& & & & \\
\midrule
number of	 	& $3\times 128$	& $3\times 128$ 	& $3\times 128$	& $3\times 128$ 	\\
hidden		& $3\times 256$	& $3\times 256$ 	& $3\times 256$	& $3\times 256$ 	\\
layers		  	& 			& 		 	&  			& \\
\midrule
number of 		& --			& -- 			& $8,12$	& $8,12$		\\
spline knots	 	& & & & \\
\midrule
total number		& $320$			& $320$			& $320$		& $400$		\\
of runs 		& & & & \\
\bottomrule
\end{tabular}
\vspace{2mm}
\caption{Hyperparameter values used in our analysis. The~last row shows the total number of runs for each architecture, with the $10$ replicas and the different~dimensionalities being taken into account.}
\label{tab:hyperparam}
\end{table}

\subsection{The~C-RQS}\label{app:CRQS}
The C-RQS parameters are determined by the following procedure~\cite{durkan2019neural}.
\ben
\item A dense DNN takes $x_{1},\ldots,x_{d}$ as inputs and~outputs an unconstrained parameter vector
$\theta_{i}$ of length $3K - 1$ for each $i = d+1,\ldots,D$ dimension. 
\item The vector $\theta_{i}$ is partitioned as $\theta_{i}=[\theta_{i}^{w},\theta_{i}^{h},\theta_{i}^{d}]$, where $\theta_{i}^{w}$ and $\theta_{i}^{h}$ have length $K$, while $\theta_{i}^{d}$ has length $K-1$.
\item The vectors $\theta_{i}^{w}$ and $\theta_{i}^{h}$ are each passed through a softmax and multiplied by $2B$; the outputs are
interpreted as the widths and heights of the $K$ bins, which must be positive and span the $\mathbb{B}$ interval. Then, the~cumulative sums of the $K$ bin widths and heights, each starting at $-B$, yield the $K+1$ knot parameters $\lbrace (x_{i}^{(k)},y_{i}^{(k)})\rbrace^{K}_{k=0}$.
\item The vector $\theta_{i}^{d}$ is passed through a softplus function and is interpreted as the values of the derivatives $\lbrace d_{i}^{(k)}\rbrace_{k=1}^{K-1}$ at the internal knots.
\een

As for the RealNVP, in~order to transform all dimensions, a~series of RQS bijectors is applied, inverting the role of the two partitions at each~step.

\subsection{The~A-RQS}\label{app:ARQS}
In the autoregressive implementation, we follow the same procedure used in the MAF implementation and described in Section~\ref{app:MAF}, but~instead of obtaining the $2(D-1)$ outputs determining the affine parameters, we obtain the $3K-1$ parameters needed to compute the values of the knots parameters and derivatives. Once these are determined, the procedure follows steps 2 to 4 of the C-RQS implementation described in the previous~subsection.

\section{Hyperparameters}\label{app:mog_hyperparam}
For all models, we used a total of $10^{5}$ trainings, $3\times 10^{4}$ validations, and~$10^{5}$ test points. We employed ReLu activation function with no regularization. All models were trained for up to $1000$ epochs, with~the \textsc{Adam} optimizer, with the initial learning rate set to $10^{-3}$.\footnote{For unstable trainings in large dimensionality, when the training with this initial learning rate failed with a ``nan'' loss, we  reduced the learning rate by a factor $1/3$ and retried until~either the training succeeded or~the learning rate was smaller than $10^{-6}$.} The learning rate was then reduced by a factor of $0.5$ after $50$ epochs without improvement better than $10^{-4}$ on the validation loss. Early stopping was used to terminate the learning after $100$ epochs without the same amount of improvement. The~batch size was set to $256$ for RealNVP and to $512$ for the other algorithms. For~the two neural spline algorithms, we also set the range of the spline equal to $[-16,16]$. The~values of all hyperparameters on which we performed a scan are reported in Table~\ref{tab:hyperparam}. 

\clearpage\markboth{}{}
\onecolumn

\section{Results Summary Tables}

\begin{table}[h!]
\footnotesize
\begin{center}
\begin{tabular}{lrlrlllrrr}
\toprule
\multicolumn{10}{c}{\bf Results for Average Best models} \\
\midrule
hidden& \# of	&  algorithm 	&  spline 	& KS 		& Sliced	& Frobenius 	& \# of  	& training 	& prediction 	\\
layers & bijec. 	&  		&  knots 	& 		 	& WD	& Norm 	&	epochs	& time (s)  	& time (s)  	\\
\midrule
\multicolumn{10}{l}{\bf 4D} \\ 
\midrule
$3\times 128$ &	10 &	MAF &	-- 	& $1.5\pm 0.4$ 	& $3.1\pm 0.7$ 	& $0.9\pm 0.3$ &	$442$ &	$5301$ &	$17$ \\
$3\times 256$ &	5 &	RealNVP &	-- 	& $1.6\pm 0.5$ 	& $3.7\pm 0.9$ 	& $1.0\pm 0.4$ &	$616$ &	$8777$ &	$8$ \\
$3\times 128$ &	2 &	A-RQS &	8 	& $1.2\pm 0.1$ 	& $2.6\pm 0.4$ 	& $0.7\pm 0.2$ &	$670$ &	$7606$ &	$54$ \\
$3\times 128$ &	5 &	C-RQS &	8 	& $1.4\pm 0.2$ 	& $3.3\pm 0.3$ 	& $1.0\pm 0.5$ &	$483$ &	$12346$ &	$26$ \\
\midrule
\multicolumn{10}{l}{\bf 8D} \\ 
\midrule
$3\times 128$ &	5 &	MAF &	-- 	& $1.3\pm 0.2$ 	& $3.3\pm 0.7$ 	& $0.8\pm 0.3$ &	$713$ &	$5757$ &	$11$ \\
$3\times 128$ &	10 &	RealNVP &	-- 	& $2.2\pm 0.5$ 	& $5.2\pm 1.1$ 	& $1.4\pm 0.4$ &	$340$ &	$8242$ &	$12$ \\
$3\times 128$ &	2 &	A-RQS &	12 	& $1.6\pm 0.5$ 	& $4.7\pm 2.1$ 	& $1.0\pm 0.4$ &	$477$ &	$5516$ &	$315$ \\
$3\times 128$ &	5 &	C-RQS &	8 	& $2.4\pm 0.6$ 	& $6.1\pm 1.9$ 	& $1.6\pm 0.5$ &	$294$ &	$11500$ &	$27$ \\
\midrule
\multicolumn{10}{l}{\bf 16D} \\ 
\midrule
$3\times 128$ &	5 &	MAF &	-- 	& $1.6\pm 0.2$ 	& $3.7\pm 0.6$ 	& $7.1\pm 1.9$ &	$479$ &	$3410$ &	$16$ \\
$3\times 128$ &	5 &	RealNVP &	-- 	& $2.6\pm 0.3$ 	& $6.4\pm 1.3$ 	& $12\pm 3$ &	$366$ &	$4263$ &	$8$ \\
$3\times 128$ &	2 &	A-RQS &	12 	& $1.8\pm 0.3$ 	& $3.1\pm 0.5$ 	& $2.2\pm 1.0$ &	$327$ &	$3705$ &	$65$ \\
$3\times 128$ &	10 &	C-RQS &	8 	& $2.0\pm 0.6$ 	& $3.9\pm 1.3$ 	& $4.6\pm 2.0$ &	$558$ &	$64056$ &	$53$ \\
\midrule
\multicolumn{10}{l}{\bf 32D} \\ 
\midrule
$3\times 128$ &	5 &	MAF &	-- 	& $1.7\pm 0.3$ 	& $4.5\pm 0.9$ 	& $2.1\pm 0.6$ &	$595$ &	$4193$ &	$24$ \\
$3\times 128$ &	10 &	RealNVP &	-- 	& $2.6\pm 0.5$ 	& $6.8\pm 1.6$ 	& $3.0\pm 0.8$ &	$676$ &	$16946$ &	$12$ \\
$3\times 128$ &	2 &	A-RQS &	8 	& $1.9\pm 0.3$ 	& $4.4\pm 1.0$ 	& $1.7\pm 0.4$ &	$375$ &	$4705$ &	$97$ \\
$3\times 256$ &	10 &	C-RQS &	12 	& $2.0\pm 0.5$ 	& $5.2\pm 1.7$ 	& $2.3\pm 0.8$ &	$750$ &	$75606$ &	$83$ \\
\midrule
\multicolumn{10}{l}{\bf 64D} \\ 
\midrule
$3\times 128$ &	10 &	MAF &	-- 	& $2.1\pm 0.6$ 	& $7\pm 3$ 	& $1.3\pm 0.5$ &	$537$ &	$5289$ &	$89$ \\
$3\times 256$ &	10 &	RealNVP &	-- 	& $2.3\pm 0.4$ 	& $6.5\pm 1.7$ 	& $1.5\pm 0.3$ &	$711$ &	$13871$ &	$16$ \\
$3\times 128$ &	2 &	A-RQS &	8 	& $1.5\pm 0.2$ 	& $4.0\pm 0.7$ 	& $0.9\pm 0.1$ &	$523$ &	$6197$ &	$332$ \\
$3\times 256$ &	5 &	C-RQS &	12 	& $2.4\pm 0.9$ 	& $6.4\pm 2.7$ 	& $2.6\pm 0.8$ &	$813$ &	$36608$ &	$53$ \\
\midrule
\multicolumn{10}{l}{\bf 100D} \\ 
\midrule
$3\times 128$ &	10 &	MAF &	-- 	& $2.2\pm 0.8$ 	& $7\pm 4$ 	& $2.1\pm 1.2$ &	$778$ &	$7824$ &	$144$ \\
$3\times 256$ &	10 &	RealNVP &	-- 	& $3.2\pm 1.3$ 	& $10\pm 6$ 	& $2.9\pm 1.9$ &	$991$ &	$23500$ &	$19$ \\
$3\times 128$ &	2 &	A-RQS &	12 	& $1.4\pm 0.2$ 	& $4.0\pm 0.9$ 	& $1.0\pm 0.2$ &	$588$ &	$6219$ &	$1027$ \\
\midrule
\multicolumn{10}{l}{\bf 200D} \\ 
\midrule
$3\times 128$ &	10 &	MAF &	-- 	& $3.7\pm 1.4$ 	& $11\pm 6$ 	& $3.4\pm 2.4$ &	$612$ &	$6149$ &	$393$ \\
$3\times 256$ &	10 &	RealNVP &	-- 	& $6.3\pm 2.3$ 	& $19\pm 11$ 	& $5.4\pm 2.2$ &	$1000$ &	$22704$ &	$25$ \\
$3\times 128$ &	2 &	A-RQS &	12 	& $1.3\pm 0.2$ 	& $3.3\pm 1.1$ 	& $1.0\pm 0.5$ &	$703$ &	$9943$ &	$4900$ \\
\midrule
\multicolumn{10}{l}{\bf 400D} \\ 
\midrule
$3\times 128$ &	10 &	MAF &	-- 	& $4.3\pm 1.8$ 	& $14\pm 8$ 	& $8\pm 6$ &	$600$ &	$4612$ &	$1242$ \\
$3\times 256$ &	10 &	RealNVP &	-- 	& $8.4\pm 2.4$ 	& $21\pm 11$ 	& $24\pm 10$ &	$824$ &	$23705$ &	$38$ \\
$3\times 128$ &	2 &	A-RQS &	8 	& $2.2\pm 0.9$ 	& $6.9\pm 1.8$ 	& $11\pm 4$ &	$796$ &	$9970$ &	$9738$ \\
\bottomrule
\end{tabular}
\end{center}
\vspace{2mm}
\caption{Values of the most relevant hyperparameters and metrics for the average best models obtained for the CMoG distributions. The number of training epochs and the training and prediction times are averages. The columns KS, Sliced WD,  and Frobenius Norm contain the values of the corresponding test statistics shown in the left panels of Figure \ref{fig:cmog_metrics}. The row corresponding to the best model for each dimension, that is, the one with the minimum KS test statistic, is shown in bold.}
\label{tab:average_best}
\end{table}

\begin{table}[h!]
\footnotesize
\begin{center}
\begin{tabular}{lrlrlllrrr}
\toprule
\multicolumn{10}{c}{\bf Results for Absolute Best models} \\
\midrule
hidden& \# of	&  algorithm 	&  spline 	& KS 		& Sliced	& Frobenius 	& \# of  	& training 	& prediction 	\\
layers & bijec. 	&  		&  knots 	& 		 	& WD	& Norm 	&	epochs	& time (s)  	& time (s)  	\\
\midrule
\multicolumn{10}{l}{\bf 4D} \\ 
\midrule
$3\times 128$ &	10 &	MAF &	-- 	& $1.1\pm 0.1$ 	& $2.0\pm 0.5$ 	& $0.5\pm 0.1$ &	$442$ &	$5301$ &	$17$ \\
$3\times 256$ &	5 &	RealNVP &	-- 	& $1.1\pm 0.2$ 	& $2.3\pm 0.6$ 	& $0.7\pm 0.3$ &	$616$ &	$8777$ &	$8$ \\
$3\times 128$ &	2 &	A-RQS &	8 	& $1.1\pm 0.2$ 	& $2.2\pm 0.6$ 	& $0.5\pm 0.2$ &	$670$ &	$7606$ &	$54$ \\
$3\times 128$ &	5 &	C-RQS &	8 	& $1.2\pm 0.2$ 	& $2.8\pm 0.7$ 	& $0.6\pm 0.2$ &	$483$ &	$12346$ &	$26$ \\
\midrule
\multicolumn{10}{l}{\bf 8D} \\ 
\midrule
$3\times 128$ &	5 &	MAF &	-- 	& $1.1\pm 0.2$ 	& $2.7\pm 0.4$ 	& $0.6\pm 0.1$ &	$713$ &	$5757$ &	$11$ \\
$3\times 128$ &	10 &	RealNVP &	-- 	& $1.3\pm 0.2$ 	& $3.2\pm 1.0$ 	& $0.6\pm 0.1$ &	$340$ &	$8242$ &	$12$ \\
$3\times 128$ &	2 &	A-RQS &	12 	& $1.1\pm 0.1$ 	& $2.6\pm 0.8$ 	& $0.6\pm 0.2$ &	$477$ &	$5516$ &	$315$ \\
$3\times 128$ &	5 &	C-RQS &	8 	& $1.4\pm 0.1$ 	& $3.1\pm 0.8$ 	& $1.0\pm 0.2$ &	$294$ &	$11500$ &	$27$ \\
\midrule
\multicolumn{10}{l}{\bf 16D} \\ 
\midrule
$3\times 128$ &	5 &	MAF &	-- 	& $1.3\pm 0.1$ 	& $2.8\pm 0.9$ 	& $2.4\pm 0.9$ &	$479$ &	$3410$ &	$16$ \\
$3\times 128$ &	5 &	RealNVP &	-- 	& $2.2\pm 0.1$ 	& $4.6\pm 0.6$ 	& $6.3\pm 0.9$ &	$366$ &	$4263$ &	$8$ \\
$3\times 128$ &	2 &	A-RQS &	12 	& $1.4\pm 0.1$ 	& $2.4\pm 0.5$ 	& $1.1\pm 0.2$ &	$327$ &	$3705$ &	$65$ \\
$3\times 128$ &	10 &	C-RQS &	8 	& $1.1\pm 0.1$ 	& $2.3\pm 0.3$ 	& $1.9\pm 0.3$ &	$558$ &	$64056$ &	$53$ \\
\midrule
\multicolumn{10}{l}{\bf 32D} \\ 
\midrule
$3\times 128$ &	5 &	MAF &	-- 	& $1.2\pm 0.1$ 	& $2.8\pm 0.7$ 	& $0.9\pm 0.2$ &	$595$ &	$4193$ &	$24$ \\
$3\times 128$ &	10 &	RealNVP &	-- 	& $1.7\pm 0.2$ 	& $4.0\pm 1.4$ 	& $1.8\pm 0.5$ &	$676$ &	$16946$ &	$12$ \\
$3\times 128$ &	2 &	A-RQS &	8 	& $1.3\pm 0.1$ 	& $2.9\pm 0.4$ 	& $1.2\pm 0.2$ &	$375$ &	$4705$ &	$97$ \\
$3\times 256$ &	10 &	C-RQS &	12 	& $1.4\pm 0.2$ 	& $2.9\pm 0.5$ 	& $1.2\pm 0.1$ &	$750$ &	$75606$ &	$83$ \\
\midrule
\multicolumn{10}{l}{\bf 64D} \\ 
\midrule
$3\times 128$ &	10 &	MAF &	-- 	& $1.5\pm 0.1$ 	& $3.1\pm 0.7$ 	& $0.8\pm 0.1$ &	$537$ &	$5289$ &	$89$ \\
$3\times 256$ &	10 &	RealNVP &	-- 	& $1.8\pm 0.1$ 	& $3.7\pm 0.8$ 	& $1.1\pm 0.2$ &	$711$ &	$13871$ &	$16$ \\
$3\times 128$ &	2 &	A-RQS &	8 	& $1.1\pm 0.2$ 	& $3.1\pm 0.8$ 	& $0.6\pm 0.1$ &	$523$ &	$6197$ &	$332$ \\
$3\times 256$ &	5 &	C-RQS &	12 	& $1.5\pm 0.1$ 	& $3.7\pm 0.5$ 	& $1.8\pm 0.1$ &	$813$ &	$36608$ &	$53$ \\
\midrule
\multicolumn{10}{l}{\bf 100D} \\ 
\midrule
$3\times 128$ &	10 &	MAF &	-- 	& $1.3\pm 0.1$ 	& $3.7\pm 0.7$ 	& $0.8\pm 0.2$ &	$778$ &	$7824$ &	$144$ \\
$3\times 256$ &	10 &	RealNVP &	-- 	& $1.9\pm 0.1$ 	& $3.1\pm 0.5$ 	& $1.4\pm 0.1$ &	$991$ &	$23500$ &	$19$ \\
$3\times 128$ &	2 &	A-RQS &	12 	& $1.1\pm 0.2$ 	& $2.8\pm 0.8$ 	& $0.8\pm 0.3$ &	$588$ &	$6219$ &	$1027$ \\
\midrule
\multicolumn{10}{l}{\bf 200D} \\ 
\midrule
$3\times 128$ &	10 &	MAF &	-- 	& $2.2\pm 0.3$ 	& $5.9\pm 1.3$ 	& $1.0\pm 0.1$ &	$612$ &	$6149$ &	$393$ \\
$3\times 256$ &	10 &	RealNVP &	-- 	& $4.0\pm 0.2$ 	& $8.8\pm 1.2$ 	& $2.9\pm 0.2$ &	$1000$ &	$22704$ &	$25$ \\
$3\times 128$ &	2 &	A-RQS &	12 	& $1.0\pm 0.1$ 	& $2.3\pm 0.7$ 	& $0.6\pm 0.1$ &	$703$ &	$9943$ &	$4900$ \\
\midrule
\multicolumn{10}{l}{\bf 400D} \\ 
\midrule
$3\times 128$ &	10 &	MAF &	-- 	& $1.7\pm 0.2$ 	& $13.2\pm 1.3$ 	& $12.0\pm 0.7$ &	$600$ &	$4612$ &	$1242$ \\
$3\times 256$ &	10 &	RealNVP &	-- 	& $5.3\pm 0.1$ 	& $10.1\pm 0.2$ 	& $8.3\pm 0.1$ &	$824$ &	$23705$ &	$38$ \\
$3\times 128$ &	2 &	A-RQS &	8 	& $1.2\pm 0.1$ 	& $3.3\pm 0.7$ 	& $2.7\pm 0.6$ &	$796$ &	$9970$ &	$9738$ \\
\bottomrule
\end{tabular}
\end{center}
\vspace{2mm}
\caption{Values of the most relevant hyperparameters and metrics for the absolute best models obtained for the CMoG distributions. The number of training epochs and the training and prediction times are averages. The columns KS, Sliced WD,  and Frobenius Norm contain the values of the corresponding test-statistics shown in the right panels of Figure \ref{fig:cmog_metrics}. The row corresponding to the best model for each dimension, that is, the one with the minimum KS test statistic, is shown in bold.}
\label{tab:absolute_best}
\end{table}

\clearpage\markboth{}{}
\onecolumn

\section{Correlation Matrix}

\begin{figure}[h]
\centering
\includegraphics[width=15.5 cm]{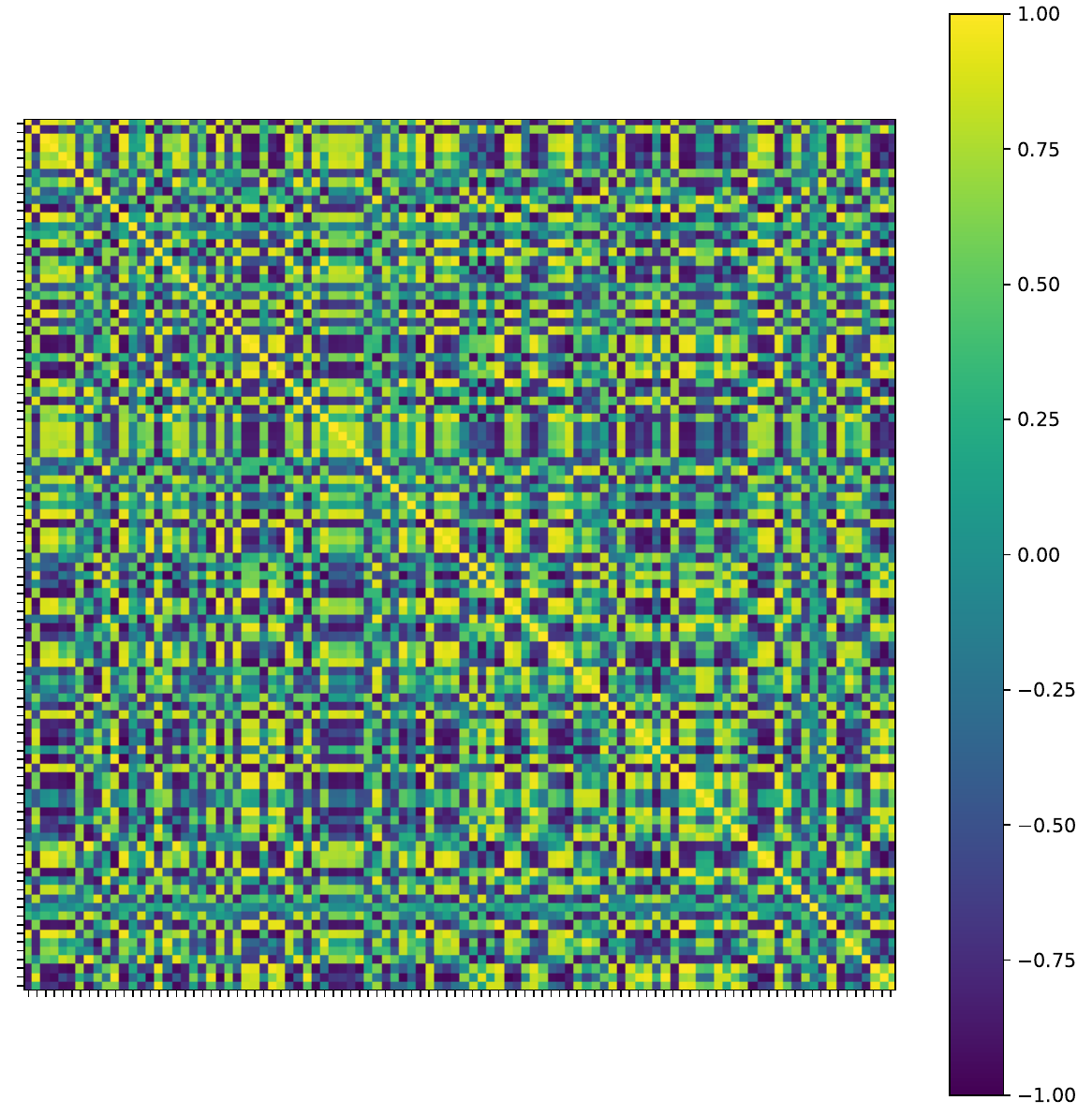}
\caption{Visual representation of the correlation matrix of our CMoG model in $N=100$ dimensions. Despite the different multivariate Gaussian components  being uncorrelated, the~resulting mixture model features random, order-one, off-diagonal elements in the full correlation matrix.}
\label{fig:corre_matrix_plot}
\end{figure}

\clearpage\markboth{}{}
\onecolumn

\newpage
\section{Corner Plots}

\begin{figure}[h]
\centering
\includegraphics[width=15.5 cm]{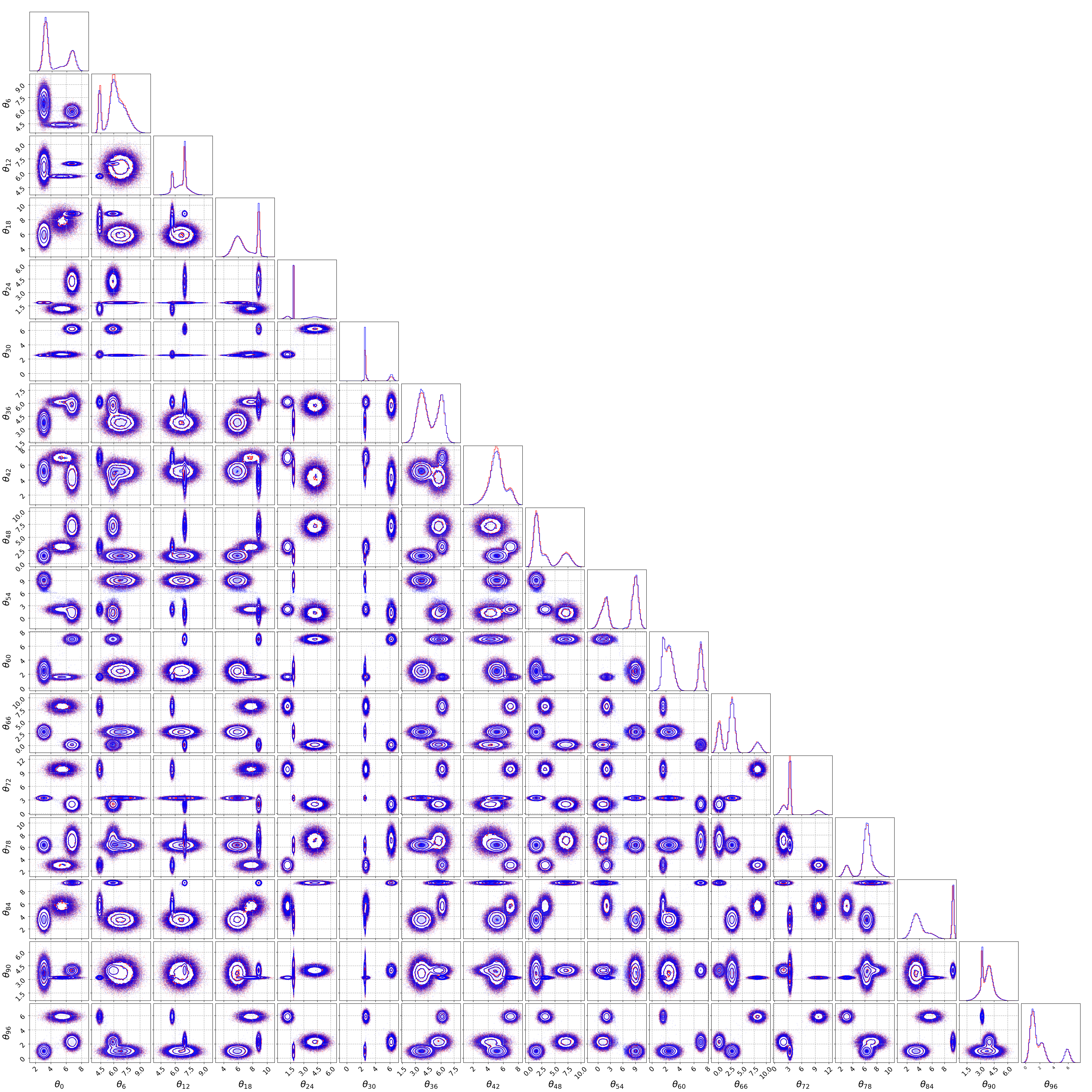}
\caption{Visual representation of the 1D and 2D marginal distributions for $17$ randomly chosen dimensions of the $N=100$ dimensional CMoG distribution obtained with $10^{5}$ points. The red and blue curves and points represent the test samples and the NF-generated samples obtained with the A-RQS best model, respectively. Given the high dimensionality, the~non-trivial structure of the distribution, the~limited number of training samples, and~the low level of tuning of the hyperparameters, the~result can be considered very~accurate.}
\label{fig:corner_corr_100}
\end{figure}

\twocolumn
\clearpage\markboth{}{}
\bibliographystyle{mine}
\bibliography{references}

\end{document}